\documentclass[11pt,letterpaper]{article}

\usepackage{apacite}
\usepackage{times}

\usepackage{graphicx}

\usepackage{color}

\usepackage[intlimits]{amsmath} 
\usepackage{bbm,amsfonts,amsthm}
\usepackage{amsmath}
\usepackage{wrapfig}

\usepackage{psfrag}

\newcommand{\tr}{\mathrm{tr\,}}

\newcommand{\argmax}{\mathop{\mathrm{argmax}}}

\newcommand{\diag}{\mathop{\mathrm{diag}}}
\newcommand{\trans}{^\text{\sffamily T}}

\newcommand{\norm}[1]{\left\|{#1}\right\|}       
\newcommand{\vect}[1]{\bigl(#1\bigr)\trans}

\newcommand{\bx}{\mathbf x}
\newcommand{\bxt}{\mathbf x_{t}}                      
\newcommand{\bxtt}{\mathbf x_{t+1}}

\newcommand{\bu}{\mathbf u}

\newcommand{\bxx}{\mathbf x'}

\newcommand{\sampleni}{\mathbf{\tilde x'}_{\nu,i}}
\newcommand{\samplenj}{\mathbf{\tilde x'}_{\nu,j}}

\newcommand{\bmu}{\boldsymbol \mu}
\newcommand{\bSigma}{\boldsymbol \Sigma}
\newcommand{\btheta}{\boldsymbol \theta}

\newcommand{\GP}{\mathcal{GP}}

\newcommand{\NMC}{N_{\text{MC}}}  

\begin{document}

\begin{titlepage}

\parindent=0pt

\begin{center}
{\huge \bf \sffamily Empowerment for Continuous Agent-Environment Systems}

\vspace{2cm}
{\Large
Department of Computer Science\\ \smallskip
The University of Texas at Austin}
\end{center}

\vspace{3cm}
{\bfseries Tobias Jung}$^1$\\
{\tt tjung@cs.utexas.edu\\}

{\bfseries Daniel Polani}$^2$\\
{\tt d.polani@herts.ac.uk\\}

{\bfseries Peter Stone}$^1$\\
{\tt pstone@cs.utexas.edu\\}

$^1$Department of Computer Science \\
University of Texas at Austin\\
1616 Guadalupe, Suite 2408\\
Austin, Texas 78701\\
USA

\bigskip

$^2$Adaptive Systems and Algorithms Research Groups\\
School of Computer Science\\
University of Hertfordshire\\
1 College Lane\\
Hatfield AL10 9AB, Herfordshire\\
United Kingdom
\end{titlepage}

\begin{center}
{\large \bf \sffamily Empowerment for Continuous Agent-Environment Systems}\\ \vspace{1.5cm}
Initial Submission September 30, 2010  \\ \vspace{0.5cm}
Revision November 4, 2010
\end{center}

\vspace{3cm}

\begin{abstract}
This paper develops generalizations of {\em empowerment} to continuous states. Empowerment is a recently introduced 
information-theoretic quantity motivated by hypotheses about the efficiency of the sensorimotor loop in biological 
organisms, but also from considerations stemming from curiosity-driven learning. 
Empowemerment measures, for agent-environment systems with stochastic transitions, how much influence an agent has on 
its environment, but only that influence that can be sensed by the agent sensors. It is an information-theoretic 
generalization of joint controllability (influence on environment) and observability (measurement by sensors) of 
the environment by the agent, both controllability and observability being usually defined in control theory as 
the dimensionality of the control/observation spaces.   
Earlier work has shown that empowerment has various interesting and relevant properties, e.g., it allows us to identify 
salient states using only the dynamics, and it can act as intrinsic reward without requiring an external reward. 
However, in this previous work empowerment was limited to the case of small-scale and discrete domains and furthermore 
state transition probabilities were assumed to be known. 
The goal of this paper is to extend empowerment to the significantly more important and relevant case of continuous 
vector-valued state spaces and initially unknown state transition probabilities. The continuous state space is addressed 
by Monte-Carlo approximation; the unknown transitions are addressed by model learning and prediction for which we apply 
Gaussian processes regression with iterated forecasting. In a number of well-known continuous control tasks we examine 
the dynamics induced by empowerment and include an application to exploration and online model learning.
\end{abstract}
\vspace*{1.5cm}


\vspace*{0.5cm}
{\bf Short title:} Empowerment for Continuous Agent-Environment Systems

\newpage

\section{Introduction}

One goal of AI research is to enable artificial agents (either virtual or physical ones) to act ``intelligently''
in complex and difficult environments. A common view is that intelligent behavior can be ``engineered''; 
either by fully hand-coding all the necessary rules into the agent, or by relying on various 
optimization-based techniques to automatically generate it. For example, in modern control and dynamic
programming a human designer specifies a performance signal which explicitly or implicitly encodes 
goals of the agent. By behaving in a way that optimizes this quantity, the agent then does
what the programmer wants it to do. For many applications, this is a perfectly reasonable approach 
that can lead to impressive results. However, it typically requires some prior knowledge and 
sometimes subtle design by the human developer to achieve sensible or desirable results.

In this paper, we investigate an approach to use the ``embodiment'' of an agent (i.e., the dynamics
of its coupling to the environment) to generate preferred behaviors without having to resort to
specialized, hand-designed solutions that vary from task to task. Our research embraces the related 
ideas of self-organization and self-regulation, where we aim for complex behavior to derive from simple 
and generic internal rules. The philosophy is that seemingly intentional and goal-driven behavior emerges 
as the by-product of the agent trying to satisfy universal rules rather than from optimizing externally 
defined rewards. Examples of this kind of work 
include {\em homeokinesis} \cite{Ay-etal08HomeostasisPI,der99:_homeok,der:self-org-robot:2000,der01:_self,zahedi10:_higher_coord_with_less_contr}, 
or the work in \cite{still-2009-85}. The second idea is 
that of intrinsically motivated behavior and artificial curiosity \cite{schmidhuber91}, where an agent engages in behavior because 
it is inherently ``interesting'' or ``enjoyable'', rather than as a step towards solving a specific (externally 
defined) goal. Intrinsically motivated behavior may not directly help in solving a goal, but there are indications that  
it leads to exploration and allows an agent to acquire a broad range of abilities which can, once the need arises, be easily
molded into goal-directed behavior. Related relevant publications include, for example, \cite{singh05intrinsicreward}.  
Other related work can be found in 
\cite{lungarella05:_method,lungarella05:_infor_self_struc,sporns:_evolv06,lungarella06:_mappin_infor_flow_sensor_networ}
and \cite{prokopenko06:_evolv_spatiot_coord_modul_robot_system,steels:04b,kaplan04:_maxim}.  

Here we will consider the principle of {\em empowerment} 
\cite{klyubin05:_all_else_being_equal_be_empow,klyubin08:_keep_your_option_open},
an information-theoretic quantity which is defined as the channel capacity between an agent's actions and its sensory 
observations in subsequent time steps. Empowerment can be regarded as ``universal utility'' which defines an a priori 
intrinsic reward or rather, a value/utility for the states in which an agent finds itself in. Empowerment is fully specified 
by the dynamics of the agent-environment coupling (namely the transition probabilities); a reward does not need to be specified. 
It was hypothesized in \cite{klyubin05:_all_else_being_equal_be_empow,klyubin08:_keep_your_option_open} 
that the greedy maximization of empowerment would direct an agent to ``interesting'' states in a variety of scenarios: 
\begin{itemize}
\item For one, empowerment can be considered a stochastic generalization of the concept of \emph{mobility} (i.e., number of
options available to an agent) which is a powerful heuristic in many deterministic and discrete puzzles and games. 
Being in a state with high empowerment gives an agent a wide choice of actions --- conversely, if an agent in 
``default mode'' poises itself a~priori in a high-empowerment state, it is best equipped to quickly move from 
there into a variety of target states in an emergency (for example, in the game of soccer, a goalkeeper who is about
to receive a penalty kick and has no prior knowledge about the player behavior to expect naturally positions himself
in the middle of the goal). In this regard the quantity of 
empowerment allows an agent to automatically (without explicit external human input) identify those states, even in complex environments. 
\item In the present paper we show that, for a certain class of continuous control problems, empowerment  
provides a natural utility function which imbues its states with an a priori value, without an explicit specification of a reward. 
Such problems are typically those where one tries to keep a system ``alive'' indefinitely, i.e., in a certain goal region for 
as long a time as possible. On the other hand, choosing the wrong actions or doing nothing would instead lead to the ``death'' 
of the system (naturally represented by zero empowerment). A natural example is pole-balancing. \footnote{Empowerment in the pole-balancing example was first investigated 
in \cite{klyubin08:_keep_your_option_open} with a discretized state space and 
{\em a priori} known state transition probabilities. Here we will strongly extend this example to the continuous case and online learning. State transition probabilities are initially not
known. Instead, the agent has to learn the transition probabilities while interacting with the environment.} In this context, we will find the 
smoothness of the system informs the local empowerment gradients around the agent's state of where the most 
``alive'' states are. Choosing actions such that the {\em local} empowerment score is maximized would then lead the agent
into those states. In the pole-balancing example this means that for a wide range of initial conditions, the agent would be made  
to balance the pendulum.  
\end{itemize}

Previous studies with empowerment showed promise in various domains but were essentially limited to the case of small-scale and finite-state
domains (the ubiquitous gridworld) and furthermore, state transition probabilities were assumed to be known a priori. The main contribution 
of this article is to extend previous work to the significantly more important case of (1) continuous vector-valued state spaces
and (2) initially unknown state transition probabilities. The first property means that we will be able to calculate
empowerment values only approximately; more specifically, here we will use Monte-Carlo approximation to evaluate the 
integral underlying the empowerment computation. The second property considers the case where the state space is previously 
unexplored and implies that the agent has to use some form of online model learning to estimate transition
probabilities from {\em state-action-successor state} triplets it encounters while interacting with the environment. Here, we will
approach model learning using Gaussian process regression with iterated forecasting.

To summarize, the paper is structured into three parts as follows:
\begin{enumerate}
\item The first part, Section~\ref{sec:informal}, gives a first, informal definition of empowerment and illustrates its general 
properties in a well-known finite-state domain.
\item The second part forms the main technical portion. Section~\ref{sect:Practical computation of empowerment} starts with 
a formal definition of empowerment for the continuous case and gives an algorithm for its computation based
on Monte-Carlo approximation of the underlying high-dimensional integrals. Section~\ref{sect:model learning} describes
model learning using Gaussian process regression (GPs) -- however, since this itself is a rather complex subject matter,
for brevity here we cannot go beyond a high-level description.
\item The third part examines empowerment empirically in a number of continuous control tasks well known in the 
area of reinforcement learning. The experiments will demonstrate how empowerment can form a natural utility measure, 
and how states with high empowerment values coincide with the natural (and intuitive) choice of a goal state in 
the respective domain. This way, if we incorporate empowerment into the perception-action loop of an agent, e.g., by 
greedily choosing actions that lead to the highest empowered states, we can obtain a seemingly goal-driven
behavior. As an application of this, we study the problem of exploration and model learning: using empowerment
to guide which parts of the state-space to exlore next, the agent can quickly ``discover the goal'' and thus 
more efficiently explore the environment -- without exhaustively sampling the state space.  
\end{enumerate}

\section{Illustrative example}
\label{sec:informal}
Although a more formal definition of empowerment will follow in the next section, here we will start by 
motivating it through a toy example. 
Informally, empowerment computes for any state of the environment the logarithm of the {\em effective} 
number of successor states the agent can induce by its actions. Thus empowerment essentially measures 
to what extent an agent can influence the environment by its actions: it is zero if, regardless what 
the agent does, the outcome will be the same. And it is maximal if every action will have a 
{\em distinct}\footnote{Meaning that for discrete state spaces, the sets of successor states are 
disjoint for differerent actions; for continuous state spaces, the domains of the underlying pdfs are non-overlapping.} outcome.
Note that empowerment is specifically  designed to allow for more general stochastic environments, of which 
deterministic transitions are just a special case. 

\begin{wrapfigure}{L}{0.17\textwidth}
\vspace{-20pt}
\begin{center}
\includegraphics[width=0.17\textwidth]{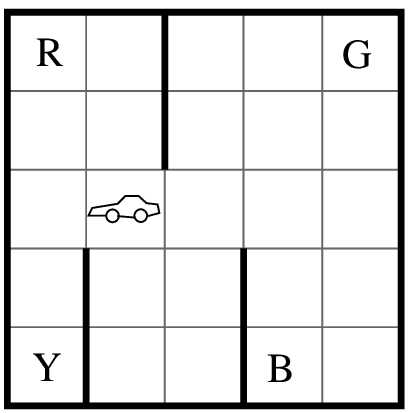}
\end{center}
\vspace{-20pt}
\end{wrapfigure}

As an example, consider the taxi-domain \cite{dietterich98taxi}, a well-known problem in reinforcement learning with finite state and action space and stochastic transitions. The environment, shown on the left, consists of a $5\times 5$ gridworld with four special locations designated 'R','Y','G','B'. Apart from the agent (``the taxi''), there is a passenger who wants to get from one of the four locations to another (selected at random). The state of the system is the $x,y$ coordinate of the agent, the location of the passenger (one of 'R','Y','G','B','in-the-car') and its destination (one of 'R','Y','G','B'). Overall there are $500=5\times 5\times 5 \times 4$ distinct states. Usually in RL, where the interest is on abstraction and hierarchical learning, a factored representation of the state is used that explicitly exploits the structure of the domain. For our purpose, where identifying salient states is part of the problem, we do not assume that the structure of the domain is known and will use a flat representation instead. The agent has six possible elementary actions: the first four ('N','S',E','W') move the agent in the indicated direction (stochastically, there is a 20\% chance for random movement). If the resulting direction is blocked by a wall, no movement occurs. The agent can also issue a pick-up and drop-off action, which require that the taxi is at the correct location and (in the latter case) the passenger is in the car. Issuing pick-up and drop-off when the conditions are not met does not result in any changes. If a passenger is successfully delivered, the environment is reset: the agent is placed in the center and a passenger with new start and destination is generated.
  
\begin{figure*}[t!]
\begin{minipage}{0.24\textwidth}
\begin{center}
\includegraphics[width=0.65\textwidth]{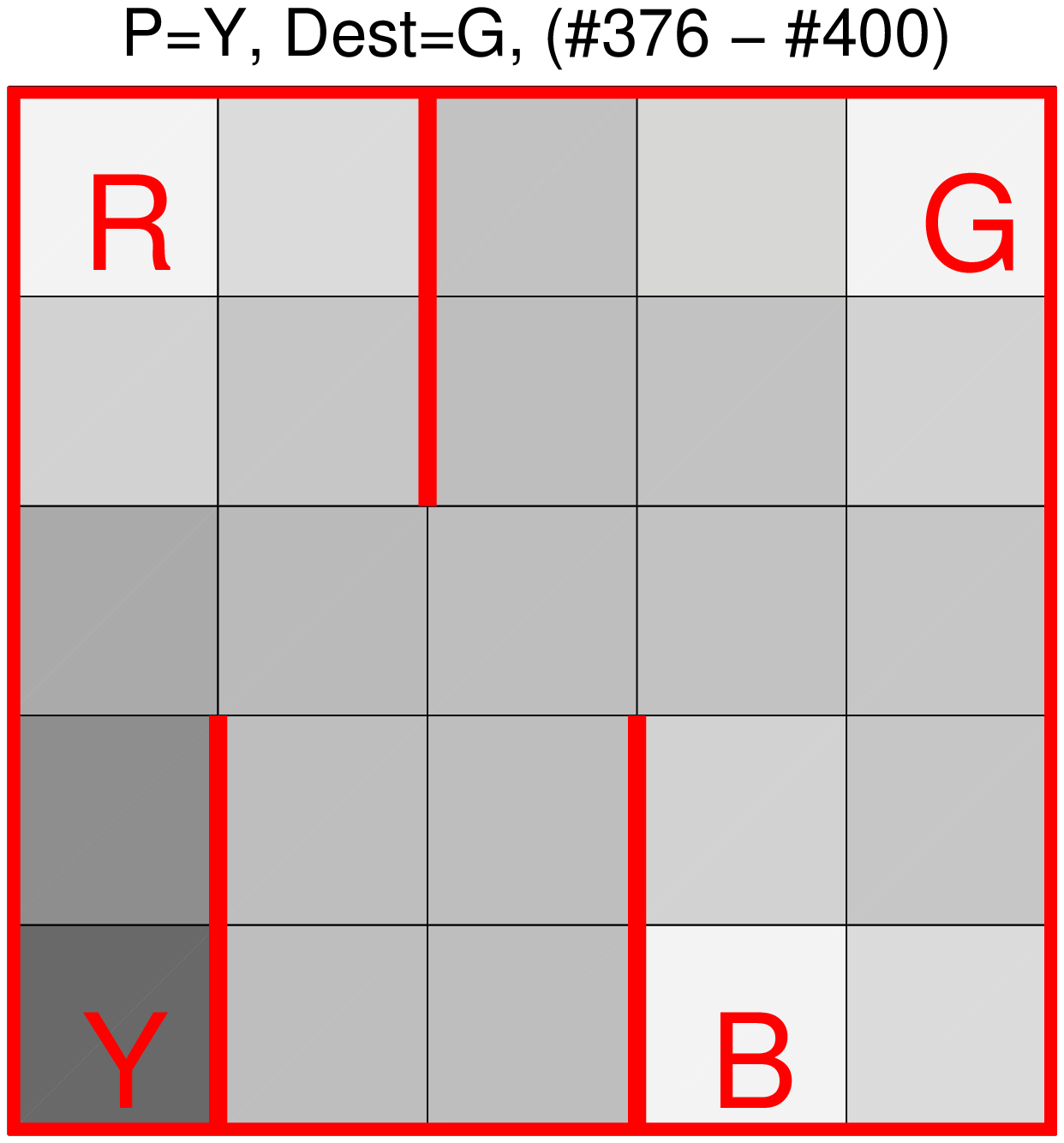}
\centerline{\small (a) P waiting at 'Y'}
\end{center}
\end{minipage}
\begin{minipage}{0.24\textwidth}
\begin{center}
\includegraphics[width=0.65\textwidth]{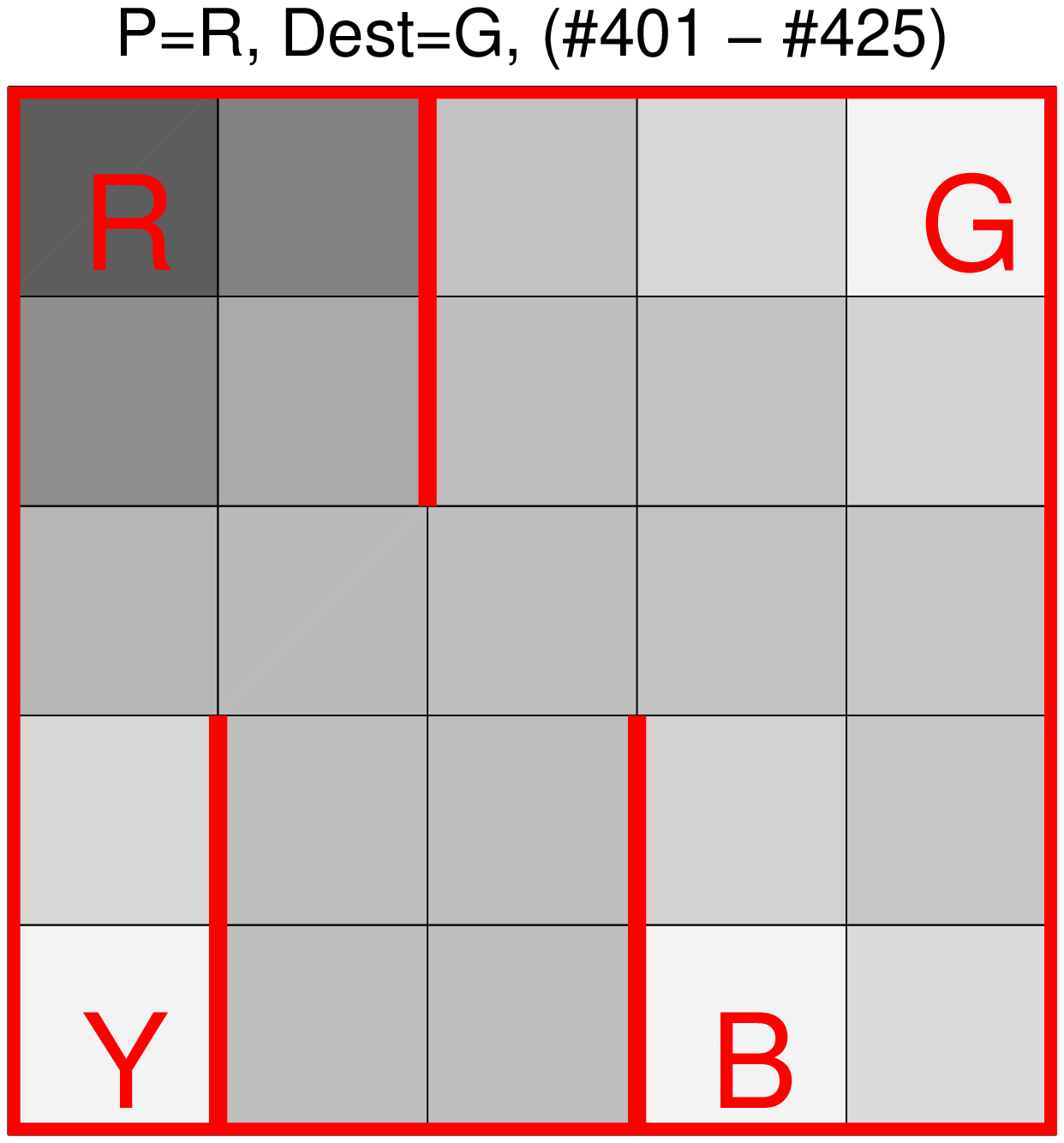}
\centerline{\small (b) P waiting at 'R'}
\end{center}
\end{minipage}
\begin{minipage}{0.24\textwidth}
\begin{center}
\includegraphics[width=0.65\textwidth]{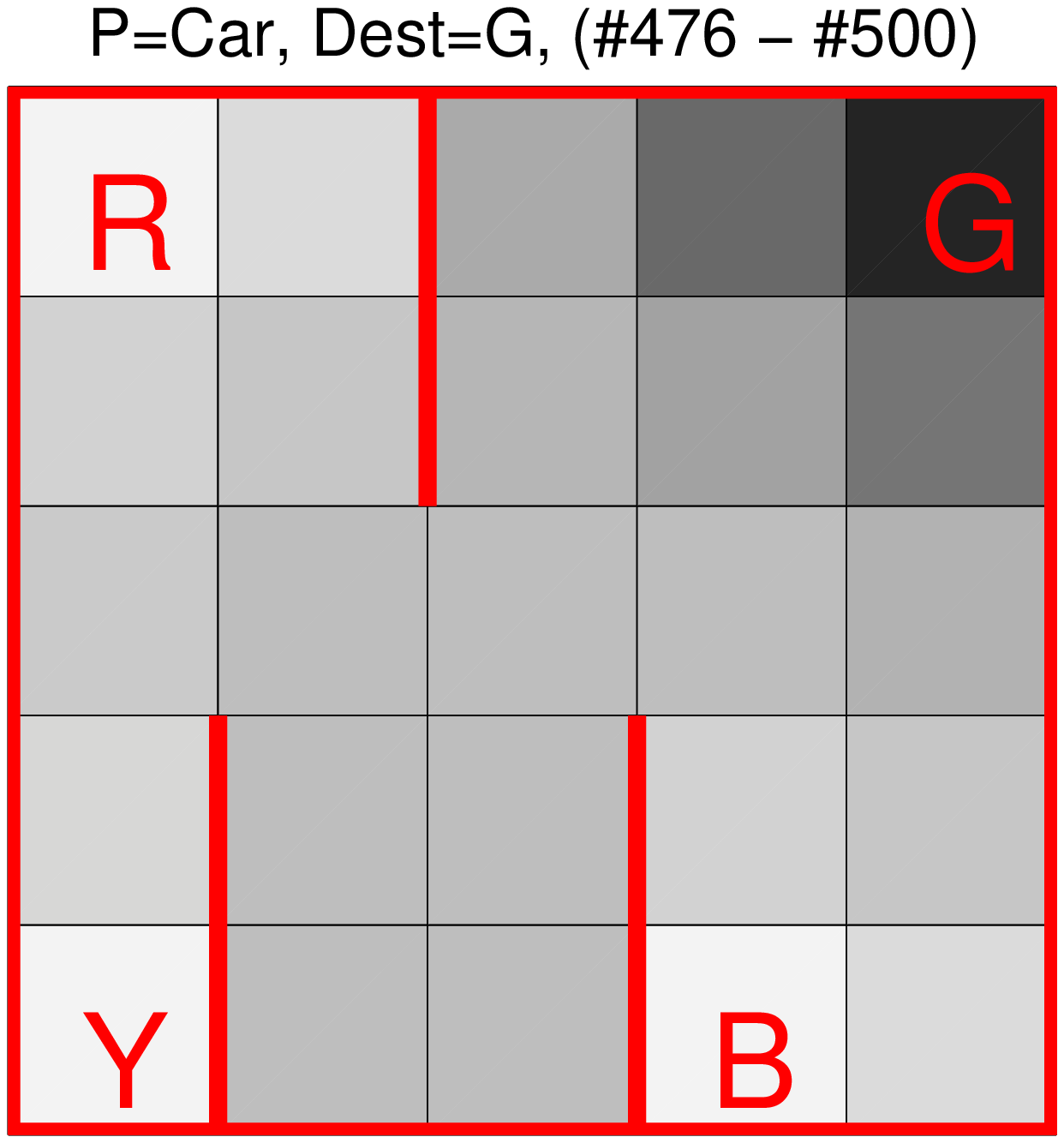}
\centerline{\small (c) P in car, going to 'G'}
\end{center}
\end{minipage}
\begin{minipage}{0.24\textwidth}
\begin{center}
\includegraphics[width=0.65\textwidth]{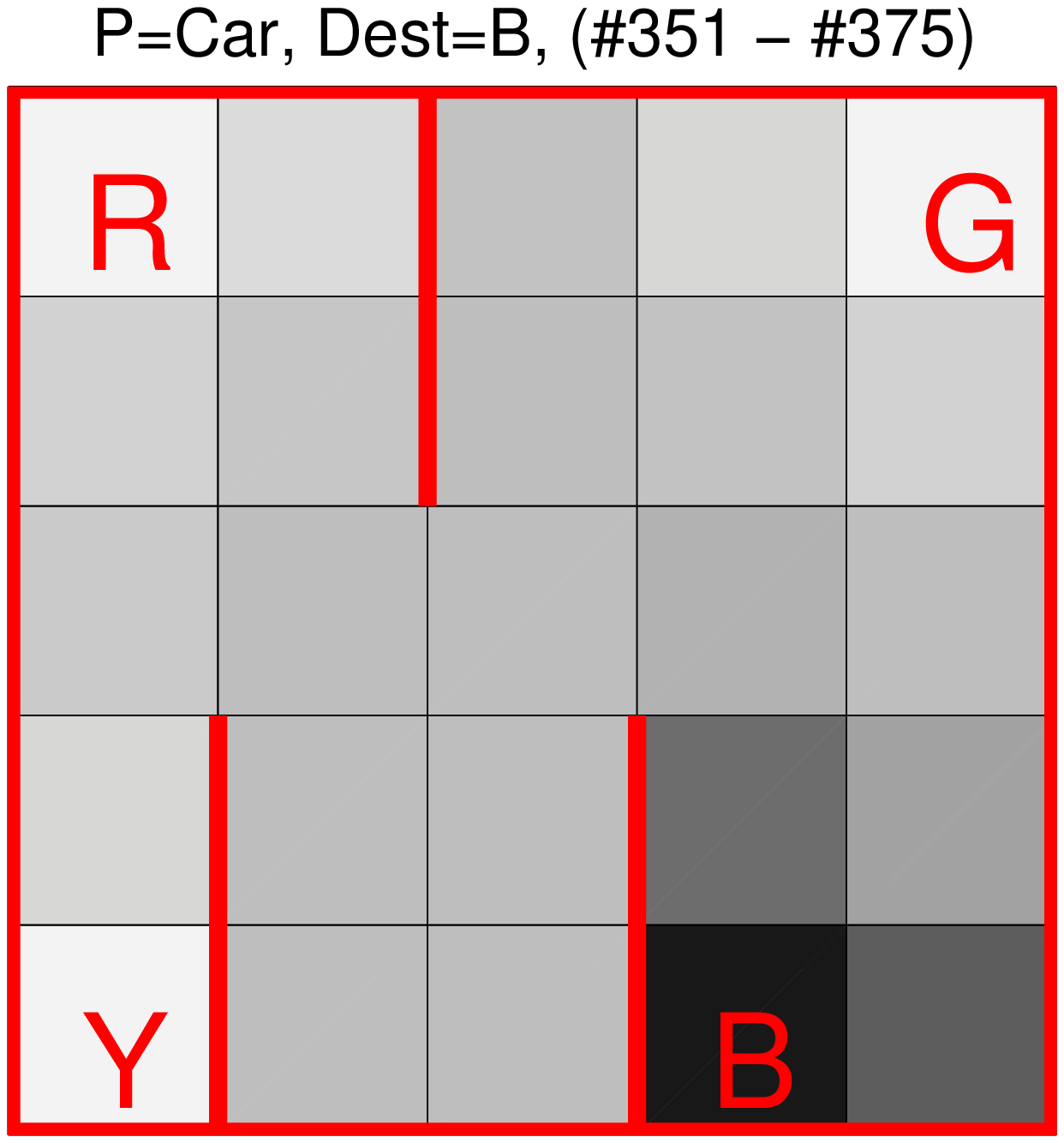}
\centerline{\small (d) P in car, going to 'B'}
\end{center}
\end{minipage}
\caption{Plotting {\em empowerment} for a subset of states (here locations) for the taxi domain. For clarity, every plot shows the mean-subtracted empowerment (3-step) of a certain slice of the state space, where white corresponds to low empowerment (1.55 nats), and black corresponds to high empowerment (2.75 nats). 
}
\label{fig:1}
\end{figure*}

Using these state transition dynamics, we compute the $3$-step empowerment, i.e., the {\em effective} number of successor states reachable over an action horizon of $3$ steps (meaning 
we consider compound actions of a sequence of three elementary actions) for every state of the system. Figure~\ref{fig:1} shows some of the results: the values are ordered such that every subplot shows the empowerment values that correspond to a specific slice of the state space. For example, the top left subplot shows the empowerment value of all $x,y$ locations if the passenger is waiting at 'Y' and its destination is 'G', which with our labeling of the states corresponds to states 376-400. Inspecting the plots, two things become apparent: for one, in general, locations in the center have high empowerment (because the agent has freedom to move wherever it wants); locations in the corners have low empowerment (because the agent has only limited choices of what it can do). More interesting is the empowerment value at the designated locations: if a passenger is waiting at a certain location, its empowerment, and that of its neighbors $2$ steps away, increases. Similarly, if a passenger is in the car, the empowerment of the destination, and that of its neighbors $2$ steps away, increases. The reason is that in both situations the agent now has additional, previously unavailable, ways of affecting the environment (plot (c) and (d) have a higher relative gain in empowerment, because they result in the end of an episode, which teleports the agent to the center). Thus these states stand out as being ``interesting'' under the heuristic of empowerment. Incidentally, these are also exactly the subgoal states if the agent's task were to transport the passenger from source to destination. Note that here we did not have to specify external reward or goals, as empowerment is intrinsically computed from the transition dynamics alone.

Empowerment essentially ``discovers'' states where additional degrees of freedom are available, and creates a basin of attraction around 
them, indicating salient features of the environment of interest to the agent. It is not difficult to imagine an agent that uses 
empowerment as a guiding principle for exploration; e.g., by choosing in each state greedily the action that leads to the successor
state with the highest empowerment. We expect that such an agent would traverse the state space in a far more sensible way than blind
random exploration, as following the trail of increasing empowerment would quickly lead to the discovery of the salient states in 
the environment. In the remainder of the paper, we will develop methods for carrying over this idea into the continuum and demonstrate
how empowerment supersedes typical hand-designed rewards in a number of established benchmark domains.

\newcommand{\sequ}{\vec{\mathbf u}_t^n}
\section{Computation of empowerment}
\label{sect:Practical computation of empowerment}
This section defines empowerment formally and gives an algorithm for its computation. 

\subsection{General definition of empowerment}
\label{sect:General definition of empowerment}
Empowerment \cite{klyubin05:_all_else_being_equal_be_empow} is defined for stochastic dynamic systems where transitions arise as the result of making a decision, e.g. such as an agent interacting with an environment. Here we will assume a vector-valued state space $\mathcal X \subset \mathbb R^D$ and (for simplicity) a discrete action space $\mathcal A=\{1,\ldots,N_A\}$. The transition function is given in terms of a density\footnote{Note that we have to consider stochastic transitions in the continuum. Otherwise if, for every action, the resulting successor states are distinct, empowerment always attains the maximum value. In practice this will usually be the case when simulating continuous control tasks with deterministic dynamics. In this case we artificially add some zero mean Gaussian noise with small variance (see Section~\ref{sec:experiment-1}). This can be interpreted as modeling limited action or sensoric resolution, depending
on the take. It is also a natural assumption for a robot realized in hardware.} $p(\bxtt|\bxt,a_t)$ which denotes the probability of going from state $\bxt$ to $\bxtt$ when making decision $a_t$. While we assume the system is fully defined in terms of these $1$-step interactions, we will also be interested in more general $n$-step interactions. Thus, for $n\ge 1$, we consider the sequence $\vec a_t^n=(a_t,\ldots,a_{t+n-1})$ of $n$ single-step actions and the induced probability density $p(\bx_{t+n}|\bx_t,\vec a_t^n)$ of making the corresponding $n$-step transition.

For notational convenience we can assume that, without loss of generality, $1$-step and $n$-step actions are equivalent: let the set of possible $n$-step actions be formed through exhaustive enumeration of all possible combinations of $1$-step actions. If $N_A$ is the number of possible $1$-step actions in every state, the number of $n$-step actions is then $N_n:=(N_A)^n$. 
With this approach, we can consider the system as evolving at the time-scale of $n$-step actions, so that $n$-step actions can be regarded as $1$-step actions at a higher level of decision making. This abstraction allows us to treat $1$-step and $n$-step actions on equal footing, which we will use to simplify the notation and drop references to the time index. Instead of writing $p(\bx_{t+n}|\bx_t,\vec a_t^n)$ we will now just write $p(\bxx|\bx,\vec a)$ to denote the transition from $\bx$ to $\bxx$ under $\vec a$, irrespective of whether $\vec a$ is an $n$-step action or $1$-step action. Furthermore we will use the symbol $\nu$ to loop over actions $\vec a$.

Let $\mathcal X'$ denote the random variable associated with $\bxx$ given $\bx$. Assume that the choice of a particular action $\vec a$ is also random and modeled by random variable $\mathcal A$. The {\em empowerment} $C(\bx)$ of a state $\bx$ (more precisely, the $n$-step empowerment) is then defined as the Shannon channel capacity (using the differential entropy) between $\mathcal A$, the choice of an action sequence, and $\mathcal X'$, the resulting successor state:
\begin{eqnarray}
C(\bx) & := & \max_{p(\vec a)} \ I(\mathcal X' ; \mathcal A \,| \, \bx) \notag \\
& = & \max_{p(\vec a)} \ \left\{ H(\mathcal X' \,| \, \bx) - H(\mathcal X' \, | \, \mathcal A, \bx) \right\}.
\label{eq:1}
\end{eqnarray}
The maximization of the mutual information is with respect to all possible distributions over $\mathcal A$, which in our case means vectors of length $N_n$ of probabilities. The entropy and conditional entropy are given by
\begin{eqnarray}
H(\mathcal X'|\bx) & := & - \int_\mathcal X p(\bxx|\bx) \log p(\bxx|\bx)d\bxx \label{eq:2a} \\
H(\mathcal X'|\mathcal A,\bx) & := & \sum_{\nu=1}^{N_n} p(\vec a_{\nu})
 H(\mathcal X'|\mathcal A=\vec a_{\nu},\bx) \notag \\
&=& -\sum_{\nu=1}^{N_n} p(\vec a_{\nu}) 
\int_\mathcal X p(\bxx|\bx,\vec a_{\nu}) \cdot \log p(\bxx|\bx,\vec a_{\nu})d\bxx. \label{eq:2b}
\end{eqnarray} 
Strictly speaking, the entropies in Eqs.~\eqref{eq:2a} and \eqref{eq:2b} are differential entropies
(which could be negative) and the probabilities are to be read as probability densities. However,
as we always end up using the mutual information, i.e. the difference between the entropies, we
end up with well-defined non-negative information values which are always finite due to the limited
resolution/noise assumed above. 
Using $p(\bxx|\bx)=\sum_{i=1}^{N_n} p(\bxx|\bx,\vec a_i)p(\vec a_i)$ in Eqs.~\eqref{eq:2a} and \eqref{eq:2b}, Eq.~\eqref{eq:1} can thus be written as
\begin{equation}
C(\bx):=\max_{p(\vec a)} \  \sum_{\nu=1}^{N_n} p(\vec a_{\nu}) \int_{\mathcal X} p(\bxx|\bx,\vec a_\nu) 
\cdot
\log 
\left\{ 
\frac{p(\bxx|\bx,\vec a_\nu)}{\sum_{i=1}^{N_n}p(\bxx|\bx,\vec a_i)p(\vec a_i)}
\right\} 
d\bxx  
\label{eq:3}
\end{equation}

Hence, given the density $p(\bxx|\bx,\vec a_{\nu})$ for making $n$-step transitions, {\em empowerment} is a function $C:\mathcal X \rightarrow \mathbbm R^{\ge 0}$ that maps an arbitrary state $\bx$ to its empowerment $C(\bx)$. 

\subsection{A concrete numerical example} 

\begin{figure}
%
%
\begin{minipage}[b]{0.49\textwidth}
\begin{center}
\includegraphics[width=0.52\textwidth]{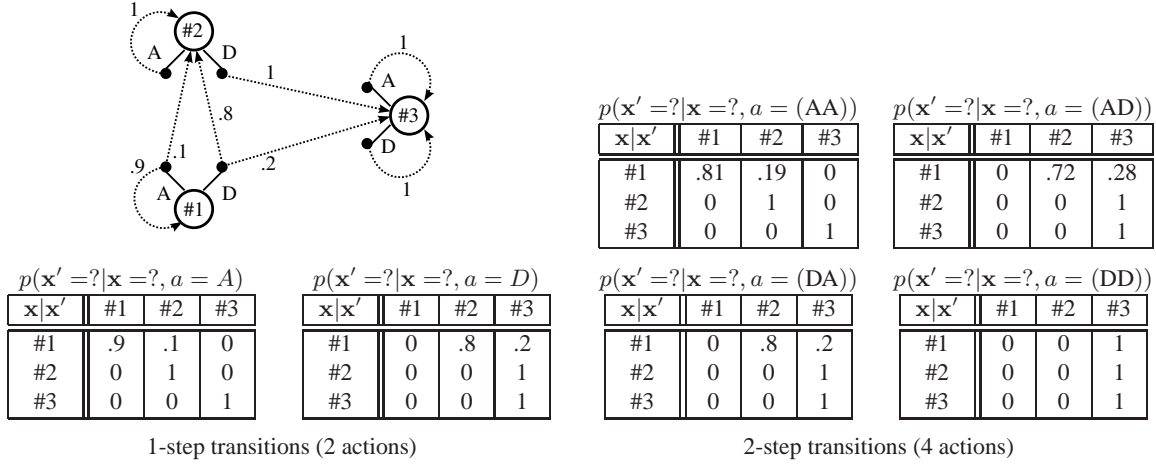}
\end{center}

\footnotesize
	\begin{minipage}{0.49\textwidth}
	\begin{center}
	$p(\bxx=?|\bx=?,a=A)$
	
	\begin{tabular}{|c||c|c|c|}
	\hline
	 $\bx | \bxx$ & \#1 & \#2 & \#3\\
	\hline 
	\hline 
	\#1 & .9 & .1 & 0 \\
	\#2 &  0 &  1 & 0 \\
	\#3 &  0 &  0 & 1 \\
	\hline
	\end{tabular}
	\end{center}
	\end{minipage}
		\begin{minipage}{0.49\textwidth}
	\begin{center}
	$p(\bxx=?|\bx=?,a=D)$
	
	\begin{tabular}{|c||c|c|c|}
	\hline
	 $\bx | \bxx$ & \#1 & \#2 & \#3\\
	\hline 
	\hline 
	\#1 &  0 & .8 & .2 \\
	\#2 &  0 &  0 &  1 \\
	\#3 &  0 &  0 &  1 \\
	\hline
	\end{tabular}
	\end{center}
	\end{minipage}
	
	\medskip
	
	\centerline{1-step transitions (2 actions)}
\end{minipage}
%
%
\begin{minipage}[b]{0.49\textwidth}
\footnotesize
	\begin{minipage}{0.49\textwidth}
	\begin{center}
	$p(\bxx=?|\bx=?,a=(\textrm{AA}))$
	
	\begin{tabular}{|c||c|c|c|}
	\hline
	 $\bx | \bxx$ & \#1 & \#2 & \#3\\
	\hline 
	\hline 
	\#1 & .81 & .19 & 0 \\
	\#2 &  0 &  1 & 0 \\
	\#3 &  0 &  0 & 1 \\
	\hline
	\end{tabular}
	\end{center}
	\end{minipage}
	\begin{minipage}{0.49\textwidth}
	\begin{center}
	$p(\bxx=?|\bx=?,a=(\textrm{AD}))$
	
	\begin{tabular}{|c||c|c|c|}
	\hline
	 $\bx | \bxx$ & \#1 & \#2 & \#3\\
	\hline 
	\hline 
	\#1 &  0 & .72 & .28 \\
	\#2 &  0 &   0 &   1 \\
	\#3 &  0 &   0 &   1 \\
	\hline
	\end{tabular}
	\end{center}
	\end{minipage}

\medskip
	\begin{minipage}{0.49\textwidth}
	\begin{center}
	$p(\bxx=?|\bx=?,a=(\textrm{DA}))$
	
	\begin{tabular}{|c||c|c|c|}
	\hline
	 $\bx | \bxx$ & \#1 & \#2 & \#3\\
	\hline 
	\hline 
	\#1 &  0 & .8 & .2 \\
	\#2 &  0 &  0 &  1 \\
	\#3 &  0 &  0 &  1 \\
	\hline
	\end{tabular}
	\end{center}
	\end{minipage}
	\begin{minipage}{0.49\textwidth}
	\begin{center}
	$p(\bxx=?|\bx=?,a=(\textrm{DD}))$
	
	\begin{tabular}{|c||c|c|c|}
	\hline
	 $\bx | \bxx$ & \#1 & \#2 & \#3\\
	\hline 
	\hline 
	\#1 &  0 &  0 & 1 \\
	\#2 &  0 &  0 & 1 \\
	\#3 &  0 &  0 & 1 \\
	\hline
	\end{tabular}
	\end{center}
	\end{minipage}	
	
	\medskip
	
	\centerline{2-step transitions (4 actions)}
	
\end{minipage}

\caption{Transition probabilities for a concrete numerical example (see text)}
\label{fig:ex2}
\end{figure}

Before we proceed, let us make the previous definition more concrete by looking at a numerical example. To simplify the exposition, 
the example will be discrete (thus integration over the domain is replaced by summation). We consider an agent in an environment 
with three states, labeled \#1,\#2,\#3, and two possible actions, 
denoted $A$ or $D$. The dynamics of the environment is fully described by the $1$-step transitions shown in 
Figure~\ref{fig:ex2}(left). The right side of the figure shows the corresponding $2$-step transitions which are derived from the 
$1$-step transitions; for example, the entry $p(\bxx=\#1|\bx=\#1,a=(AA))$ is obtained by
\begin{eqnarray*}
p(\bxx=\#1|\bx=\#1,a=(AA))&=&\sum_{i=\#1}^{\#3} p(\bxx=\#1|\bx=i,a=A)\cdot p(\bxx=i|\bx=1,a=A) \\
&=& (.9 \times .9) + (.1 \times 0) + (0 \times 0)=.81 .
\end{eqnarray*}

Let us now assume we want to calculate the $2$-step empowerment value $C(\#1)$ for state $\bx=\#1$. First, consider the $2$-step
mutual information, $I(\mathcal X';\mathcal A|\bx=\#1)$, for state $\bx=\#1$. According to Eq.~\eqref{eq:1}, we have
\begin{eqnarray*}
I(\mathcal X';\mathcal A|\bx=\#1)&=& 
p(AA)\cdot \sum_{i=\#1}^{\#3}p(\bxx=i|\bx=\#1,a=AA) 
\log \left\{ \frac{p(\bxx=i|\bx=\#1,a=AA)}{p(\bxx=i|\bx=\#1)} \right\} \\
&+& p(AD)\cdot \sum_{i=\#1}^{\#3}p(\bxx=i|\bx=\#1,a=AD) 
\log \left\{ \frac{p(\bxx=i|\bx=\#1,a=AD)}{p(\bxx=i|\bx=\#1)} \right\} \\
&+& p(DA)\cdot \sum_{i=\#1}^{\#3}p(\bxx=i|\bx=\#1,a=DA) 
\log \left\{ \frac{p(\bxx=i|\bx=\#1,a=DA)}{p(\bxx=i|\bx=\#1)} \right\} \\
&+& p(DD)\cdot \sum_{i=\#1}^{\#3}p(\bxx=i|\bx=\#1,a=DD) 
\log \left\{ \frac{p(\bxx=i|\bx=\#1,a=DD)}{p(\bxx=i|\bx=\#1)} \right\}.
\end{eqnarray*} 
The denominator in the logarithm is calculated for any $i$ via:
\begin{eqnarray*}
p(\bxx=i|\bx=\#1) 
&=& p(\bxx=i|\bx=\#1,a=AA) \cdot p(AA) \\
&+& p(\bxx=i|\bx=\#1,a=AD) \cdot p(AD) \\
&+& p(\bxx=i|\bx=\#1,a=DA) \cdot p(DA) \\
&+& p(\bxx=i|\bx=\#1,a=DD) \cdot p(DD) 
\end{eqnarray*}

As we can see, the resulting value for $I(\mathcal X';\mathcal A|\bx=\#1)$ will only depend on the individual 
probabilities of the actions, $p(AA),p(AD),p(DA),p(DD)$, but not on the transition probabilities as these are
fixed for a given environment. One natural choice for the action probabilities could be the uniform distribution.
However, for empowerment we try to find an assignment of action probabilities such that the resulting
$I(\mathcal X';\mathcal A)$ value is maximimized among all possible assignments (an algorithm for this will
be given in the next section). Below we have calculated the empowerment values MI (taking uniform distribution
over actions) and Em (taking the maximizing distribution over actions) in our example for various time horizons, 
i.e., $1$-step, $2$-step, etc. Note that, while empowerment values are logarithmic, for the purpose of illustration the results are given in terms of $\exp(I(\mathcal X';\mathcal A))$:

\begin{center}
\begin{tabular}{|c||cc|cc|cc|cc|cc|}
\hline
& \multicolumn{2}{|c|}{1-step} & \multicolumn{2}{|c|}{2-step} & \multicolumn{2}{|c|}{3-step} &
  \multicolumn{2}{|c|}{4-step} & \multicolumn{2}{|c|}{5-step}\\
\hline
State & MI & Em & MI & Em & MI & Em & MI & Em & MI & Em \\
\hline \hline
$\bx=\#1$ & 1.70 & 1.71 & 1.93 & 2.17 & 1.81 & 2.10  & 1.58 & 2.05 & 1.38 & 2.02 \\
$\bx=\#2$ & 2    & 2    & 1.75 & 2    & 1.45 & 2     & 1.26 & 2    & 1.14 & 2 \\
$\bx=\#3$ & 1    & 1    & 1    & 1    & 1    & 1     & 1    & 1    & 1    & 1 \\
\hline
\end{tabular}
\end{center}

The first column, $1$-step, illustrates the full range of possible empowerment values. Empowerment in state \#3 is
zero (here, $1=\exp(0)$), because all actions in \#3 have the same outcome. Empowerment in state \#2 is maximal
(here $2$, corresponding to the two possible $1$-step actions), because each action in \#2 has a different outcome.
In state \#1 the set of successor states overlap, thus the empowerment value is in between the two extremes.

As the time horizon increases, we can make the following observations. One is that the empowerment value
of \#3 always stays at zero, because no matter what the agent does, the outcome will be the same (thus absorbing
states are ``dead'' states). Two, the MI value of \#2 goes down, whereas its Em value stays constant (this in fact
is an important observation). The reason is that, as the time horizon increases, so does the number of 
possible ($n$-step) actions, e.g., $32=2^5$ for $5$ steps. However, a large number of these actions will bring 
the agent into \#3 from which it cannot escape. Therefore, if all actions contribute in equal parts to the result (which they 
do in MI, where we assume a uniform distribution), those that lead to zero empowerment will dominate and thus also 
the end result will be close to zero. On the other hand, the maximization in Em will suppress the effect
of indistinguishable actions (assigning zero probability to actions having the same outcome and high probabilities to actions 
having distinct outcomes) and thus ensure that the two distinct choices in \#2 are always correctly identified.

\subsection{Empowerment or mutual information?} 
Let us summarize. Empowerment measures to what extent an agent can influence the environment by 
its actions. It specifically works for stochastic systems (where state transitions are given
in terms of probabilities), but can also apply to deterministic systems (which are just a special case 
of stochastic systems). Empowerment is zero if, regardless what the agent does, the outcome will be the same 
(i.e., the outcome distribution for a given successor state $\bx'$ is independent of the action). And it 
is maximal if every action will have a distinct outcome (i.e., the probability that a single outcome
is produced by two different actions is zero). 

Let us now briefly discuss why the related information-theoretic quantity mutual information, which would 
largely have the same properties and would be easier to compute, is not as powerful as channel capacity at 
identifying interesting states of the environment. 

\begin{wrapfigure}{L}{0.14\textwidth}
\vspace*{-20pt}
\psfrag{a1}{\tiny $a_1$}
\psfrag{a2}{\tiny $a_2$}
\psfrag{a00}{\tiny $a_{99}$}
\psfrag{a01}{\tiny $a_{100}$}
\begin{center}
\includegraphics[width=0.14\textwidth]{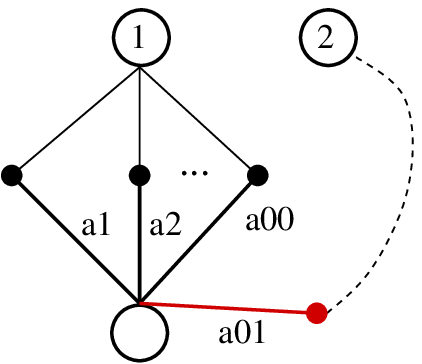}
\end{center}
\vspace*{-20pt}
\end{wrapfigure}

First, let us comment that to use the idea of modeling the influence of the action channel, one has to define some kind of distribution on the actions. As
we are considering only an agent's embodiment, but have not defined a controller, there is no default action distribution
that one could use. Therefore, one has to distinguish particular action distributions for which the action channel is to be measured. The main natural choices are the choice of an action distribution that is equally distributed,
not singling out any particular action, and that one which maximizes $I(\mathcal X';\mathcal A)$, i.e. the one that achieves
channel capacity. As we have seen in the last section, the equidistribution of actions can fail to resolve important properties
of the action channel which the optimal distribution does detect.
The most obvious situation is one where one has a large number of equivalent actions. If mutual information assumes a uniform
distribution over actions, it will be mislead by large numbers of actions that lead to the same outcome. 
As another example, consider the following situation. Assume an agent 
has $100$ different actions available and is in a state where every action has the same effect (empowerment 
and mutual information both zero). Now let us assume the agent enters a new state, as shown on the left side, where actions $a_1$ to $a_{99}$ still have the same outcome (state 1), but one action $a_{100}$ leads to a different state (state 2). In this case, use of mutual information 
with equidistributed would still be close to zero ($\approx 0.05$ nats), indicating that all actions roughly have the same effect, 
whereas empowerment correctly identifies two distinct choices ($\approx0.69=\log(2)$ nats) since it will redistribute the actions in
a way that highlights the additional degrees of freedom attained by $a_{100}$.

\subsection{Computing empowerment when a model is available}
\label{sect:Computing empowerment when a model is available}
Next we describe the Blahut-Arimoto algorithm for computing the channel capacity given in Eq.~\eqref{eq:3}. 
For now we assume that the ($n$-step) transition probabilities $p(\bxx|\bx,\vec a_\nu)$ are known for 
all actions $\vec a_\nu, \nu=1,\ldots,N_n$.

\subsubsection{Blahut-Arimoto algorithm}
\label{sect:Blahut-Arimoto algorithm}
The Blahut-Arimoto algorithm \cite{blahut72} is an EM-like algorithm that iterates over distributions $p_k(\vec a)$, where $k$ denotes the $k$-th iteration step, to produce the distribution $p^*(\vec a)$ that achieves the maximum in Eq.~\eqref{eq:3}. Since we consider a discrete action domain, $p_k(\vec a)$ is represented by a vector $p_k(\vec a) \equiv \bigl( p_k^1,\ldots,p_k^{N_n} \bigr)$. To avoid cluttered notation, we define
\begin{equation}
d_{\nu,k}:=\int_\mathcal{X} p(\bxx|\bx,\vec a_\nu) \log \left[
\frac{p(\bxx|\bx,\vec a_\nu)}{\sum_{i=1}^{N_n} p(\bxx|\bx,\vec a_i)p_{k}^i}
\right]
d\bxx.
\label{eq:7}
\end{equation}

We start with an initial distribution $p_0(\vec a)$ which is chosen using the uniform distribution,
that is $p_0^\nu:=1/N_n$ for $\nu=1,\ldots,N_n$. At each iteration $k\ge 1$, the probability
distribution $p_k(\vec a)$ is then obtained from $p_{k-1}(\vec a)$ as
\begin{equation}
p_k^\nu:=z_k^{-1} p_{k-1}^\nu \exp (d_{\nu,k-1}) \quad \nu=1,\ldots,N_n
\label{eq:4}
\end{equation}
where $z_k$ is a normalization ensuring that the new probabilities sum to one, i.e.
\begin{equation}
z_k=\sum_{\nu=1}^{N_n} p_{k-1}^\nu \,
\exp(d_{\nu,k-1}).
\label{eq:5}
\end{equation}
Once $p_k(\vec a) \equiv \bigl( p_k^1,\ldots,p_k^{N_n} \bigr)$ is computed for iteration $k$,
we can use it to obtain an estimate $C_k(\bx)$ for the empowerment $C(\bx)$ given in Eq.~\eqref{eq:3}
via
\begin{equation}
C_k(\bx)=\sum_{\nu=1}^{N_n} p_k^\nu \, \cdot \, d_{\nu,k}.
\label{eq:6}
\end{equation}
The algorithm in Eqs.~\eqref{eq:4}-\eqref{eq:6} can either be carried out for a fixed number of iterations, or it can be stopped once the change $|C_k(\bx) - C_{k-1}(\bx)|<\varepsilon$ drops below a chosen threshold and hence $C_k(\bx)$ is reasonably close to $C(\bx)$.

One problem still remains, which is the evaluation of the high-dimensional
integral over the state space in $d_{\nu,k}$.

\subsubsection{Monte-Carlo integration}
\label{sect:Monte-Carlo integration}
Taking a closer look at Eq.~\eqref{eq:7}, we note that $d_{\nu,k}$ can also be written as expectation
with regard to the density $p(\bxx|\bx,\vec a_\nu)$. 
Assuming that each density $p(\bxx|\bx,\vec a_\nu)$ is of a simple form (e.g. parametric,
like a Gaussian or a mixture of Gaussians) from which we can easily draw $\NMC$ samples
$\{\sampleni\}$, we have  
\begin{equation}
\forall \nu: \quad 
d_{\nu,k}
\approx 
\frac{1}{\NMC} \sum_{j=1}^{NMC} \log \left[
\frac{p(\samplenj|\bx,\vec a_\nu)}{\sum_{i=1}^{N_n} p(\samplenj|\bx,\vec a_i)p_{k}^i}
\right]
\label{eq:9}
\end{equation}
%

\subsubsection{Example: Gaussian model}
\label{sec:Example: Gaussian model}
As an example consider the case where $p(\bxx|\bx,\vec a_\nu)$ is a multivariate Gaussian (or at least reasonably well approximated by it) with known mean vector $\bmu_\nu=\vect{\mu_{\nu,1},\ldots,\mu_{\nu,D}}$
and covariance matrix $\bSigma_\nu=\diag\bigl( \sigma^2_{\nu,1},\ldots,\sigma^2_{\nu,D}\bigr)$, which in short will be written as
\begin{equation}
\bxx|\bx,\vec a_\nu \sim \mathcal N(\bmu_\nu, \bSigma_\nu).
\label{eq:10a}
\end{equation}
Note that here both the mean and covariance will depend on the action $\vec a_\nu$ and the state $\bx$. 
Samples $\mathbf{\tilde x'}_\nu$ from Eq.~\eqref{eq:10a} are easily generated via standard algorithms.

In summary, to compute the empowerment $C(\bx)$ given state $\bx \in \mathcal X$ and transition model 
$p(\bxx|\bx,\vec a_\nu)$, we proceed as follows.
\begin{enumerate}
\item {\bf Input:}
   \begin{enumerate}
      \item State $\bx$ whose empowerment we wish to calculate.
      \item For every action $\nu=1,\ldots,N_n$ a state transition model $p(\bxx|\bx,\vec a_\nu)$, 
      each fully defined by its mean $\bmu_\nu$ and covariance $\bSigma_\nu$.
   \end{enumerate}
\item {\bf Initialize:} 
   \begin{enumerate}
     \item $p_0(\vec a_\nu):=1/N_n$ for $\nu=1,\ldots,N_n.$
     \item Draw $\NMC$ samples $\sampleni$ each, from $p(\bxx|\bx,\vec a_\nu)=\mathcal
            N(\bmu_\nu,\bSigma_\nu)$ for $\nu=1,\ldots,N_n$. 
     \item Evaluate $p(\sampleni|\bx,\vec a_\mu)$ for all $\nu=1,\ldots,N_n$; $\mu=1,\ldots,N_n$; $i=1,\ldots,\NMC$.
   \end{enumerate}
\item {\bf Iterate} $k=1,2,\ldots$ (until $|c_k-c_{k-1}|<\texttt{tol}$ or maximum number of iterations reached) 
  \begin{enumerate}
  \item $z_k:=0$, $c_{k-1}:=0$
  \item For $\nu=1,\ldots,N_n$
        \begin{enumerate}
        \item  $d_{\nu,k-1}:=$
        \[
           \frac{1}{\NMC} \sum_{j=1}^{\NMC} \log \left[
               \frac{p(\samplenj|\bx,\vec a_\nu)}{\sum_{i=1}^{N_n} p(\samplenj|\bx,\vec a_i)p_{k-1}(\vec a_i)}
               \right]
        \]
        \item $c_{k-1}:=c_{k-1}+p_{k-1}(\vec a_\nu)\cdot d_{\nu,k-1}$
        \item $p_k(\vec a_\nu):=p_{k-1}(\vec a_\nu) \cdot \exp\{d_{\nu,k-1}\}$
        \item $z_k:=z_k+p_k(\vec a_\nu)$
        \end{enumerate} 
  \item For $\nu=1,\ldots,N_n$
        \begin{enumerate}
        \item $p_k(\vec a_\nu):=p_k(\vec a_\nu)\cdot z_k^{-1}$
        \end{enumerate}
  \end{enumerate}
\item {\bf Output:}
   \begin{enumerate}
      \item Empowerment $C(\bx)\approx c_{k-1}$ (estimated).
      \item Distribution $p(\vec a)\approx p_{k-1}(\vec a)$ achieving the maximum mutual information. 
   \end{enumerate}
\end{enumerate}
At the end we obtain the estimated empowerment $C_{k-1}(\bx)$ from $c_{k-1}$ with associated distribution
$p_{k-1}(\vec a)\equiv\bigl(p_{k-1}(\vec a_1),\ldots,p_{k-1}(\vec a_{N_n})\bigr)$. The computational cost of this algorithm is $\mathcal O(N_n^2 \cdot \NMC)$ operations per iteration; the memory requirement is $\mathcal O(N_n^2 \cdot \NMC)$. Thus the overall computational complexity scales with the square of the number of ($n$-step) actions $N_n$.

\section{Model learning}
\label{sect:model learning}
In this section we further reduce our assumptions, and consider an environment for which neither $n$-step nor 1-step transition probabilities are readily available. Instead, we assume that we could only observe a number of 1-step transitions which are given as triplets of state, performed action, and resulting successor state. Using regression on these samples, we first infer a 1-step transition model. Proceeding from this 1-step model we can then obtain a more general $n$-step transition model through iteratively predicting $n$ steps ahead in time.

In general, there would be many ways the task of regression could be accomplished. Here we will use Gaussian process regression (GP) \cite{raswil06gp}. GPs are simple and mathematically elegant, yet very powerful tools that offer some considerable advantages. One is that GPs directly produce a predictive distribution over the target values, which is exactly what is needed in Eq.~\eqref{eq:3} for the computation of empowerment. Furthermore, the predictive distribution is Gaussian and hence easy to draw samples from during the Monte-Carlo approximation (see Section~\ref{sec:Example: Gaussian model}). Also, GPs are non-parametric, meaning that a GP model is not restricted to a certain class of functions (such as polynomials), but instead encompasses {\em all} functions sharing the same degree of smoothness. In practice GPs are also very easy to use: the solution can be found analytically and in closed form. The Bayesian framework allows us to nicely address the problem of hyperparameter selection in a principled way, which makes the process of using GPs virtually fully automated, i.e. without having to adjust a single parameter by hand.

\subsection{Learning 1-step system dynamics}
\label{sect:Learning 1-step system dynamics}
To learn the state transition probabilities $p(\bx'|\bx,a=\nu)$, i.e. predict the successor state $\bx'$ when performing $1$-step action $a=\nu$ in state $\bx$, we combine multiple univariate GPs. Each individual $\GP_{\nu j}$, where $j=1\ldots D$ and $\nu=1\ldots N_A$, predicts the $j$-th coordinate of successor state $\bxx$ under action $a=\nu$. Each individual $\GP_{\nu j}$ is trained independently on the subset of the transitions where action $\nu$ was chosen: the desired target outputs we regress on is the change in the state variables (i.e. we predict the difference $\bx_{t+1}-\bx_t$). Since both state variables and actions are treated separately, we need a total of $D\cdot N_A$ independent GPs.

\begin{figure}

\psfrag{s1}{$x_1$}
\psfrag{s2}{$x_2$}
\psfrag{sd}{$x_D$}
\psfrag{GPi1}{$\GP_{\nu 1}$}
\psfrag{GPi2}{$\GP_{\nu 2}$}
\psfrag{GPid}{$\GP_{\nu D}$}
\psfrag{s}{$\bx$}
\psfrag{ss}{$\bxx$}
\psfrag{ss1}{$x'_1$}
\psfrag{ss2}{$x'_2$}
\psfrag{ssd}{$x'_D$}

\psfrag{Nss1}{$\mathcal N\bigl(\mu_{\nu 1}(\bx),\sigma_{\nu 1}^2(\bx)\bigr)$}
\psfrag{Nss2}{$\mathcal N\bigl(\mu_{\nu 2}(\bx),\sigma_{\nu 2}^2(\bx)\bigr)$}
\psfrag{Nssd}{$\mathcal N\bigl(\mu_{\nu D}(\bx),\sigma_{\nu D}^2(\bx)\bigr)$}

\psfrag{Input:}{{\bf Input:} state $\bx=\vect{x_1,\ldots,x_D}$, action $a=\nu$}
\psfrag{Output:}{{\bf Output:} $p(\bxx|\bx,a=\nu)=
\mathcal N\bigl(\bmu_\nu(\bx),\bSigma_\nu(\bx)\bigr)$
}

\psfrag{Nss}{$\mathcal N\bigl(\bmu_\nu(\bx),\bSigma_\nu(\bx)\bigr)$}

\begin{center}
\includegraphics[width=0.7\textwidth]{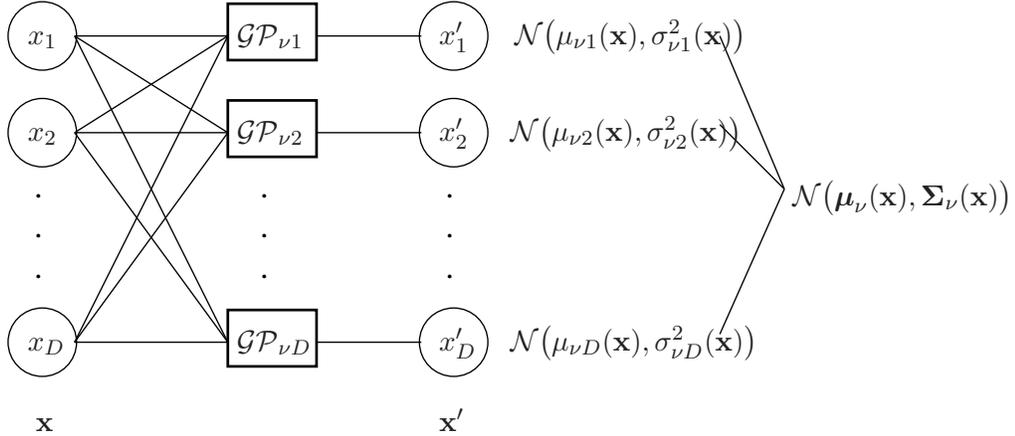}
\end{center}
\caption{Learning state transition probabilities $p(\bxx|\bx,a=\nu)$ by combining
multiple univariate GPs. Each individual $\GP_{\nu j}$ predicts the $j$-th coordinate of
successor state $\bxx$ under action $a=\nu$. Each individual $\GP_{\nu j}$ is trained independently
on the corresponding subset of the training data 
and has its own set of hyperparameters $\btheta_{\nu j}$ (obtained from maximizing the marginal
likelihood).}
\label{fig:gp}
\end{figure}

A detailed description of how univariate regression with GPs work\footnote{There is also the problem of 
implementing GPs {\em efficiently} when dealing with a possible large number of data points. For brevity 
we will only sketch our particular implementation, see \cite{jqc07approxpg} for more detailed information. 
Our GP implementation is based on the {\em subset of regressors} approximation. The elements of the subset are 
chosen by a stepwise greedy procedure aimed at minimizing
the error incurred from using a low rank approximation (incomplete Cholesky decomposition). Optimization of the
likelihood is done on random subsets of the data of fixed size. To avoid a degenerate predictive variance,
the {\em projected process} approximation was used.} can be found in \cite{raswil06gp}. Training $\GP_{\nu j}$ gives us a distribution $p(x'_j|\bx,a=\nu)=\mathcal N(\mu_{\nu j}(\bx),\sigma^2_{\nu j}(\bx))$ for the $j$-th variable of the successor state, where the exact equations for the mean $\mu_{\nu j}(\bx)$ and variance  $\sigma^2_{\nu j}(\bx)$ can be found in \cite{raswil06gp}. Note that every $\GP_{\nu j}$ will have its own set of hyperparameters $\btheta_{\nu j}$, each independently obtained from the associated training data via Bayesian hyperparameter selection. Combining the predictive models for all $D$ variables, we obtain the desired distribution 
\begin{equation}
p(\bx'|\bx,a=\nu)=\mathcal N(\bmu_\nu(\bx),\bSigma_\nu(\bx))
\label{eq:10b}
\end{equation}
for making a 1-step transition from $\bx$ under action $a=\nu$, where 
$\bmu_\nu(\bx)=\vect{\mu_{\nu 1}(\bx),\ldots,\mu_{\nu D}(\bx)}$,
and $\bSigma_\nu(\bx)=\diag \bigl(\sigma_{\nu 1}^2(\bx),\ldots,\sigma_{\nu D}^2(\bx)\bigr)$.
See Figure~\ref{fig:gp} for an illustration of this situation.

\subsection{From 1-step to $n$-step models}
\label{sect:From 1-step to n-step models}
To turn the 1-step model into an $n$-step model $p(\bx_{t+n}|\bx_t,\vec a_t^n)$, where $\vec a_t^n=(a_t,a_{t+1},\ldots,a_{t+n-1})$ is a sequence of $n$ 1-step actions, we have to integrate over all intermediate distributions. Unfortunately, solving this integral analytically in closed form is not possible.
%
%
One simple approach is again to use sampling methods, like the Monte-Carlo approximation, to numerically determine the integral. Alternatively, one could consider a more sophisticated approximate solution based on the Laplace approximation, as was proposed in \cite{girard03noisyinputs}.

\newcommand{\bxhat}{\mathbf{\hat x}}

Since, in our experiments, we will only consider very short prediction horizons (typically $n=3$ or $n=5$), we will use the more naive approach
of predicting iteratively $n$ steps ahead using the learned 1-step model. Given state $\bx_t$, we apply Eq.~\eqref{eq:10b} to produce 
$p(\bx_{t+1}|\bx_t, a_t)$. Instead of considering the full distribution, we just take its mean $\bxhat_{t+1}:=\bmu_{a_{t}}(\bx_t)$ as point estimate 
and use that to predict $\bx_{t+2}$,  applying again the 1-step model Eq.~\eqref{eq:10b} to produce $p(\bx_{t+2}|\bxhat_{t+1}, a_{t+1})$. Repeating 
this procedure until the end of the prediction horizon is reached, we obtain after $n$ steps $p(\bx_{t+n}|\bxhat_{t+n-1},a_{t+n-1})$ as an 
approximation to the originally sought $n$-step transition model $p(\bx_{t+n}|\bx_t,\vec a_t^n)$. In general, this approximation will tend to 
underestimate the variance of the prediction and produce a slightly different mean, since every time we produce an estimate for $t+i$, we 
ignore the uncertainty in the preceding prediction for $t+i-1$. In our case, however, the procedure will incur only a neglible error since 
the prediction horizon we consider is very short. See \cite{girard03noisyinputs} for more details.

\section{Experiments}
\label{sect:experiments}

We have indicated earlier that empowerment has shown intuitively appealing identification of salient states in discrete scenarios
and we are now ready to study a number of more intricate continuous scenarios. These scenarios are used as benchmark
for typical learning algorithms (e.g., reinforcement learning or optimal control). However, it should be noted that
in the latter the learning algorithms need to be instructed about which optimization criterion to use in the learning 
process. Here, we will always use empowerment maximization as the criterion, and demonstrate that the resulting
behaviors actually match closely those where optimization of an external quality criterion is requested. The observation
that these behaviors match, is a subtle point and will be discussed in more detail in the discussion (see Section~\ref{sec:discussion}).

As an important side effect, empowerment can also be used as a (heuristic) exploration driver in these 
scenarios: this is particularly interesting since, unlike optimal control algorithms, empowerment
is fundamentally local (limited to the horizon defined by the actions) as opposed to optimal control algorithms
that, for an informed decision, need to have their horizon extended to encompass information about 
the desired target state(s) to a sufficiently accurate extent.

Thus, in the following section, we will demonstrate that 
\begin{enumerate}
\item empowerment {\em alone} can lead to apparently intentional and goal-directed behavior of an agent based only on the embodiment of the agent with no external reward structure, and 
\item how it can furthermore act as a heuristic to guide the agent's exploration of the environment. 
\end{enumerate}
We  consider two scenarios: one {\em without model learning}, and one {\em with model learning}. The first scenario 
will demonstrate that incorporating empowerment into the perception-action loop of an agent produces intuitively 
desirable behavior, by greedily choosing actions in each state that lead to the highest empowered states. Our primary intent 
here is to show that empowerment itself is a relevant quantity to be considered and for simplicity we assume that the 
transition probabilities of the system are known. In the second scenario, we will further reduce our assumptions and
consider this no longer to be the case. The agent starts out knowing nothing about the environment it is in. We will then combine empowerment with 
model learning and exploration: while, as in the first scenario, the agent chooses its actions based on empowerment, the underlying 
computations are carried out using a {\em learned} model for the state transition probabilities. The model is 
continually updated (in batches) from the transitions the agent experiences and thus gets continually better at 
predicting the effects the actions will have, which in turn will produce more accurate empowerment values. 
A comparison with common model-based reinforcement learning, RMAX \cite{RMAX-JMLR02}, which operates in
a similar fashion but actively optimizes an external performance criterion, concludes.

\subsection{The domains}
As testbeds for our experiments, we consider simulations of the three physical systems described below. 
We reiterate that, in the literature, systems like these are usually used in the context of control and learning behavior where 
a goal (desired target states) is {\em externally} defined and, by optimizing a thus determined performance criterion, the system 
is driven to specifically reach that goal. In contrast, empowerment used here is a {\em generic} heuristic (aimed at curiosity-driven
learning) where a goal is not explicitly defined and which operates on innate characteristics of the system's dynamic alone. 
It will turn out that empowerment intrinsically drives the system (close) to states which in fact are typically externally chosen 
as goal states. However, with empowerment we do not enforce this goal through any external reward but through a generic intrinsic
quantity that, for each domain, is generated in exactly the same way. Note that, in a wider sense, all the tasks belong to the class of 
control problems where the goal is to choose actions such that the system stays ``alive'' -- to achieve this, the agent has to 
stay in a certain ``stable'' goal region. This is a class of problems for which we believe empowerment is particularly well-suited.

\begin{figure}
\begin{minipage}{0.32\textwidth}
\psfrag{e}{\small$\dot\omega$}
\psfrag{w}{\small$\omega$}
\psfrag{a}{\small$\phi$}
\psfrag{b}{\small$\dot\phi$}
\includegraphics[scale=0.7]{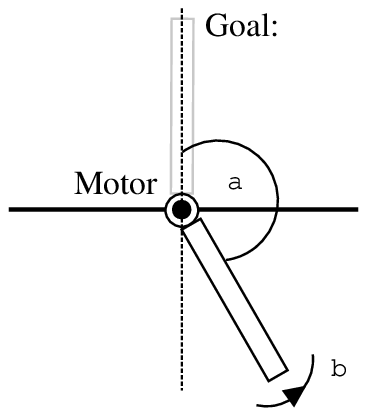} \bigskip \\
\includegraphics[scale=0.8]{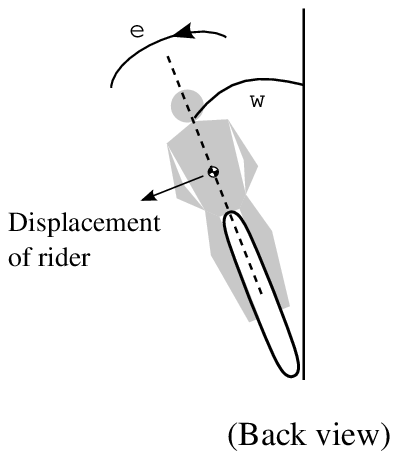}
\end{minipage}
\begin{minipage}{0.32\textwidth}
\psfrag{r}{\small$r$}
\psfrag{m}{\small$M_c$}
\psfrag{n}{\small$M_r$}
\psfrag{o}{\small$M_d$}
\psfrag{l}{\small$l$}
\psfrag{h}{\small$d_{CM}$}
\includegraphics[scale=0.8]{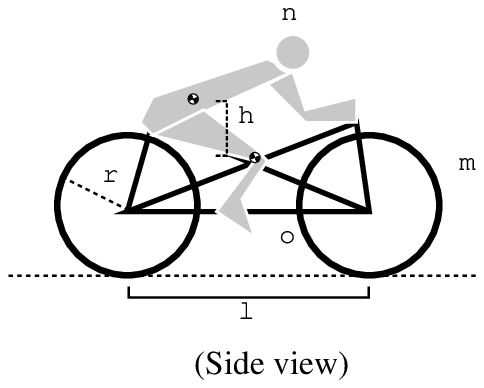} \bigskip \\
\psfrag{b}{\small$\alpha$}
\psfrag{c}{\small$\dot\alpha$}
\includegraphics[scale=0.8]{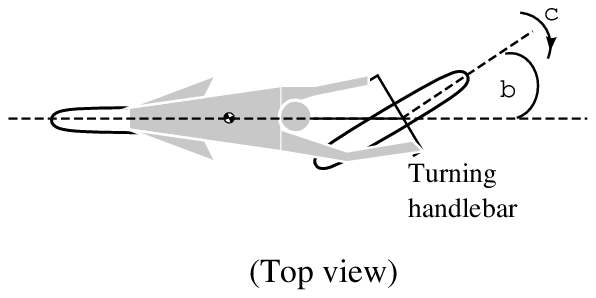}
\end{minipage}
\begin{minipage}{0.32\textwidth}
\psfrag{a}{\small$\theta_1$}
\psfrag{b}{\small$\theta_2$}
\psfrag{c}{\small$\dot\theta_1$}
\psfrag{d}{\small$\dot\theta_2$}
\includegraphics[scale=0.7]{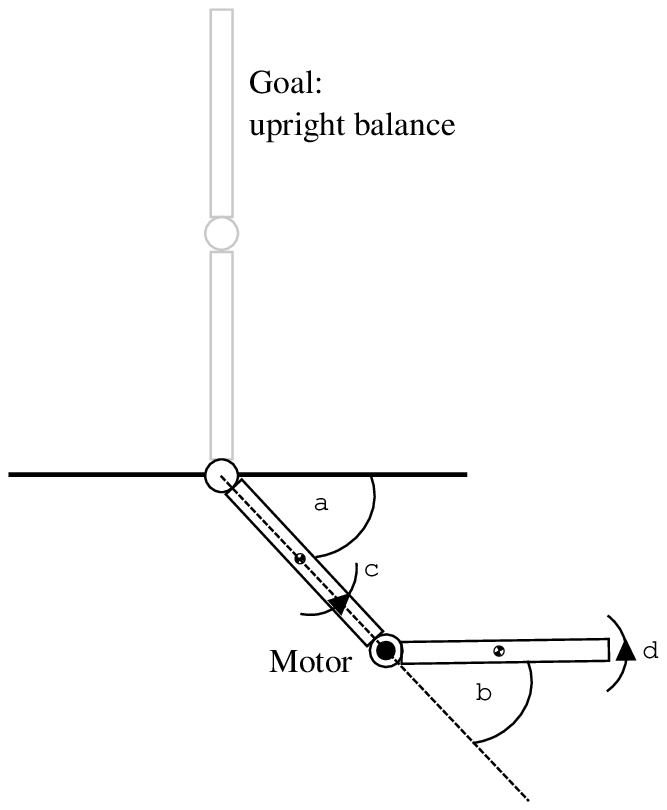}
\end{minipage}
\caption{From left to right: the inverted pendulum task, the riding a bicycle task, and the acrobot handstand task.}
\label{fig:domains}
\end{figure}

\paragraph{Inverted pendulum:} The first system consists of a single pole attached at one end to a motor, as depicted in Figure~\ref{fig:domains}. 
If force is applied, the pole will freely swing in the $xy$ plane. More detailed dynamic equations of the system are given in 
the appendix. If no force is applied, the system returns to its stable equilibrium (pole hangs down vertically). Let this state be the 
initial condition. The goal is to swing up and stabilize the pole in the inverted position. However, the motor does not provide enough torque to do so directly 
in a single rotation. Instead, the pendulum needs to be swung back and forth to gather energy, before being pushed up and
balanced. This creates a somewhat difficult, nonlinear control problem. The state space is $2$-dimensional, $\phi \in [-\pi,\pi]$ being the angle,
$\dot\phi \in [-10,10]$ the angular velocity. Since our empowerment model only deals with a finite number of $1$-step and $n$-step actions, 
the control force is discretized to $a\in \{-5,-0.25,0,0,+0.25,+0.5\}$. 

\paragraph{Riding a bicycle:} The second domain is a more involved one: we consider the bicycle riding 
task described in \cite{Lagoudakis03,Ernst05} and depicted in Figure~\ref{fig:domains}. In this task, a bicycle-rider system (modeled as 
a simplified mechanical system) moves at a constant speed 
on a horizontal surface. The bicycle is not self-stabilizing and has to be actively stabilized to be prevented from falling. The goal is to 
keep the bicycle stable such that it continues to move forward indefinitely. A detailed description of the dynamics of the system is given
in the appendix. The problem is $4$-dimensional: state variables are the roll angle $\omega \in [-12\pi/180,12\pi/180]$, roll rate 
$\dot\omega \in [-2\pi,2\pi]$, angle of the handlebar $\alpha \in [-80\pi/180,80\pi/180]$, and the angular velocity 
$\dot\alpha \in[-2\pi,2\pi]$. The control space is inherently $2$-dimensional: $u_1$, the horizontal displacement of the bicycle-rider system 
from the vertical plane, and $u_2$, turning the handlebar from the neutral position. Since empowerment
can only deal with a finite number of $1$-step and $n$-step actions, we consider $5$ possible action vectors: 
$(u_1,u_2) \in \{(-0.02,0), (0,0), (0.02,0), (0,-2),(0,2)\}$.

\paragraph{Acrobot:} The third domain is the acrobot proposed in \cite{Spong95}. The acrobot can be imagined as a gymnast 
swinging up above a high bar by bending at the hips. As depicted in Figure~\ref{fig:domains}, the acrobot is a two-link robot, which freely swings
around the first joint (the hands grasping the bar) and can exert force only at the second joint (the hips). Controlling the acrobot 
is a very challenging problem in nonlinear control; it is underactuated, meaning that the dimensionality of the state space is higher
than that of the actuators, or, informally, that it has more degrees of freedom than actuators 
(in robotics, many systems are underactuated, including manipulator arms on spacecraft, non-rigid body systems, and balancing systems
such as dynamically stable legged robots). Usually two tasks are considered for the acrobot in the literature: the first and easier 
one is to swing the tip (the feet) of the lower link over the bar at the height of the upper link. The second task is significantly 
more difficult: as in the first task, the goal is to swing up the lower link; however, this time the acrobot has to reach the inverted handstand 
position with close to zero velocity, and then to actively balance so as to remain in this highly unstable state for as long as possible. 
A detailed description of the dynamics of the system is given in the appendix. The initial state of the acrobot is the stable equilibrium 
with both links hanging vertically down. The state space is $4$-dimensional: $\theta_1 \in [-\pi,\pi]$, $\dot\theta_1 \in [-4\pi,4\pi]$,
$\theta_2 \in [-\pi,\pi]$, $\dot\theta_2 \in [-9\pi,9\pi]$. Since, as before, empowerment can deal with only a finite number of $1$-step
and $n$-step actions, the continuous control was discretized to $a \in \{-1,+1\}$. However, while these two actions alone are sufficient to 
solve the swing-up task, they are not sufficient for the inverted balance, since for this case, control values between the two extremes $-1$ 
and $+1$ must be chosen. Therefore, we include a third, non-primitive 'balance' action, which chooses control values derived from an LQR controller
obtained from linearizing the system dynamics about the handstand position (see appendix). Note that this 'balance' action produces meaningful 
(i.e., actually useful) outputs only very close to the handstand state which means that it cannot be naively used to direct the acrobot to 
balance from an arbitrary point of the state space.

\subsection{First scenario: model-based}
\label{sec:experiment-1}
In our first series of experiments,
the agent chooses actions greedily to maximize empowerment. For all domains, we assume that the 
state transition probabilities are known. 
The control loop becomes the following: every time 
step $t$ the agent observes the current state $\bx_t$. Using the state transition function, we 
determine the $1$-step successor states under each of the possible $1$-step actions. For each of 
these states, we compute the empowerment value as described in Section~\ref{sec:Example: Gaussian model}, 
using $\NMC=200$, $\texttt{TOL}=10^{-5}$ and $\texttt{MAX\_ITER}=150$, and adding Gaussian white 
noise with (state-independent) covariance to ``smear out'' the otherwise deterministic state 
transitions. The agent then executes the action corresponding to the successor state with the highest 
empowerment value (empowerment-greedy action selection), advancing the time and producing the next 
state $\bx_{t+1}$. 

Note that in practice, for empowerment values to be meaningful, we usually require an increased 
look-ahead horizon into the future than just a single simulation step; thus, instead of $1$-step empowerment, we usually 
need to examine $n$-step empowerment for values of $n$ greater than one. Here we form the $n$-step actions 
through exhaustive enumeration; thus if $N_A$ is the number of possible $1$-actions the agent 
has available, the number $N_n$ of $n$-step actions we have to consider during the computation of empowerment 
is $N_n=(N_A)^{n}$. 
For each experiment performed, we informally\footnote{Note that the metaparameters time horizon, 
simulation step size (at what frequency the controls are allowed to change), and what amount of noise to add 
are not independent from each other and must be chosen judiciously and for each domain seperately. If, for example, 
the variance of the noise is too small relative to the average distance of the successor states (which depends 
on the horizon), then empowerment will always be close to maximal (the mass of the distributions does not ``overlap'' 
and all successor states are treated as distinct). On the other hand, if the noise is too large relative to the average distance of the 
successor states, then empowerment can no longer distinguish between the effects of different actions (because 
the individual results are ``smeared out`` too much). At this time we do not have a full analytical understanding 
of how the parameters interact and how to best determine them for a given domain in a disciplined way 
other than by trial and error.} determined the minimum time horizon of lookahead necessary 
to achieve the desired effect. Especially for small simulation steps (such as $\Delta=0.01$), the number
$n$ of $1$-step actions needed to fill a given time horizon could grow relatively large, which in turn 
would then lead to a large number of $n$-step actions, rendering computational costs prohibitive. To 
reduce the number of $n$-step actions while still maintaining the same lookahead, each $1$-step action 
in an action sequence was held constant for an extended amount of time, a multiple of the 
simulation step $\Delta$. An alternative would be to intelligently compress and prune the lookahead tree, as 
suggested in \cite{anthony09:_impov_empow} for discrete scenarios, which there allows to extend the horizon
by more than an order of magnitude at similar complexity. Here, however, we are going to demonstrate that
even the locally informed empowerment with short lookahead horizons is sufficient to treat aforementioned 
scenarios.

\begin{figure}[ht]
\begin{minipage}{1.09\textwidth}
\begin{center}
\hspace*{-0.9cm}\includegraphics[width=0.327\textwidth]{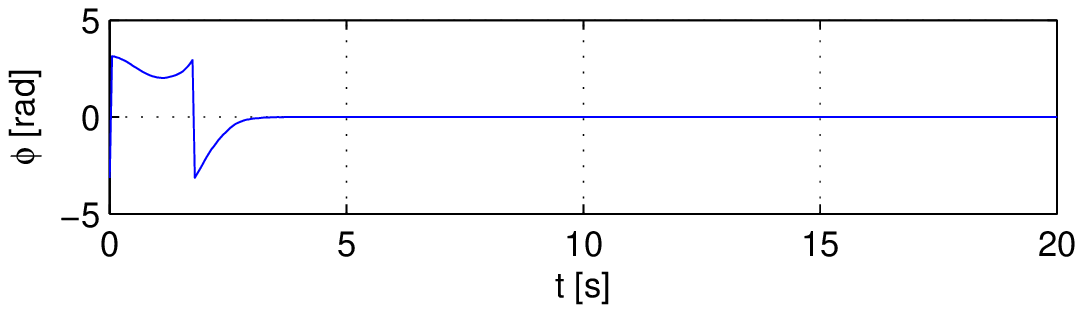} \hspace*{-0.7cm}
\includegraphics[width=0.327\textwidth]{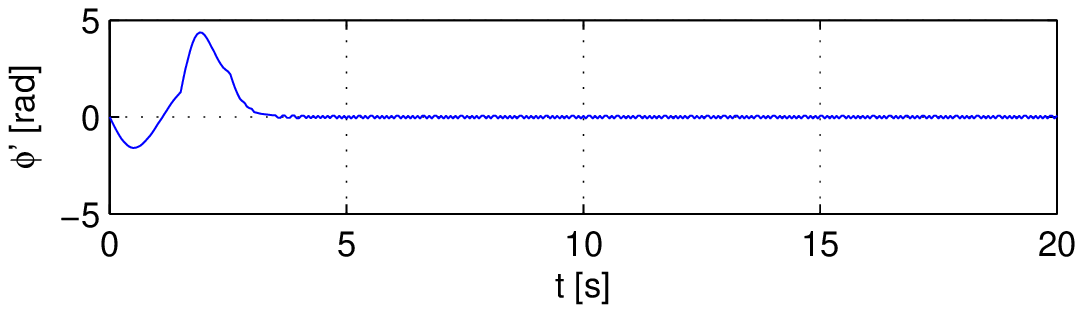} \hspace*{-0.7cm}
\includegraphics[width=0.327\textwidth]{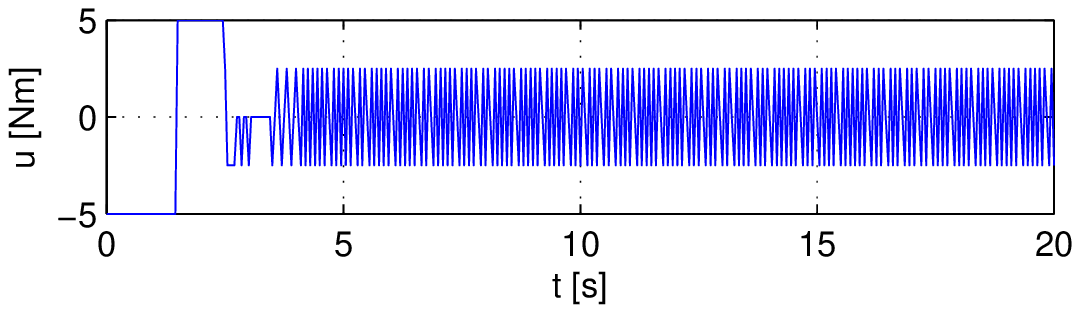}
\end{center}
\end{minipage}

\medskip

\begin{minipage}{1.09\textwidth}
\begin{center}
\hspace*{-0.9cm}\includegraphics[width=0.327\textwidth]{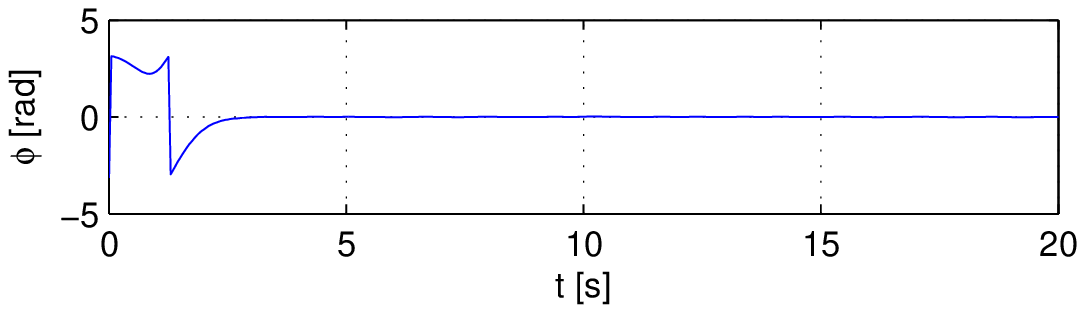} \hspace*{-0.7cm}
\includegraphics[width=0.327\textwidth]{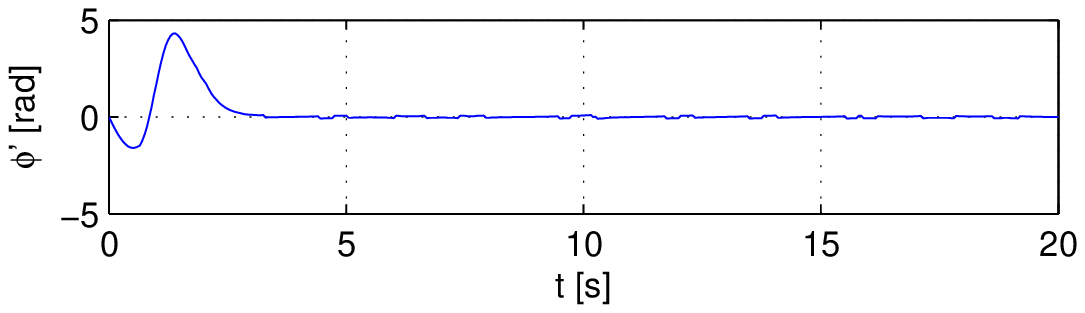} \hspace*{-0.7cm}
\includegraphics[width=0.327\textwidth]{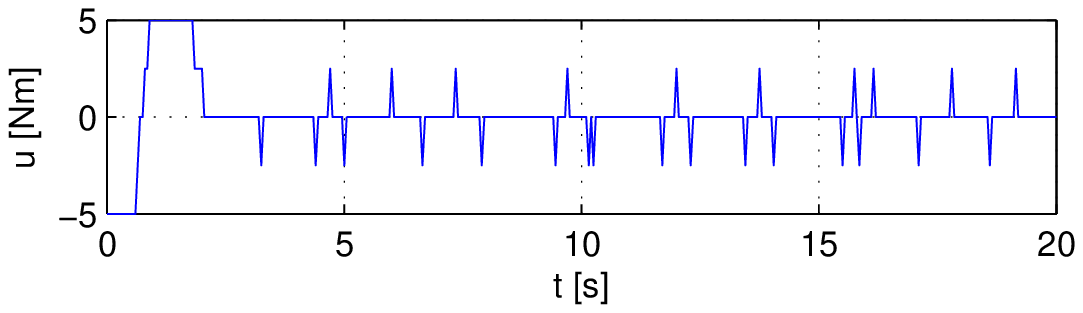}
\end{center}
\end{minipage}
\caption{Inverted pendulum: phase plot of $\phi, \dot\phi$ and control $u$ when following the greedy policy with respect to: empowerment (top row); dynamic programming (bottom row). 
}
\label{fig:phaseplot_pendulum}
\end{figure}

\paragraph{Results for inverted pendulum:}
Figure~\ref{fig:phaseplot_pendulum} (top row) shows a phase plot of the behavior that results from starting 
in the initial condition (pole hanging vertically down) and following $3$-step empowerment 
(and thus $N_n=5 \times 5 \times 5$ $n$-step actions) for a period of 20 seconds with state transition 
noise $\bSigma=0.01 \mathbf{I}_{2\times 2}$ (where $\mathbf{I}_{n\times n}$ denotes the $n\times n$ 
identity matrix).   
The plot demonstrates that: (1) empowerment alone makes the agent drive up the pendulum and successfully 
balance it indefinitely; (2) the agent accomplishes the goal without being explicitly ``told'' to do so; 
and (3) the trajectory shows that this happens in a straight and direct way, without wasting time (and 
consistently so). Note that empowerment only ``illuminates'' the local potential future of the current state
and has no access to the global value of the trajectory as opposed to optimal control methods where 
implicitly global information about the goal states must be propagated back throughout the system model
for the controller to take the right decision.  

To compare these results with a different angle, we reformulate the problem as a minimum-time optimal control 
task: as opposed to before, we now assume that the agent has an explicit, externally specified goal (swinging up 
the pendulum as fast as possible and successfully balancing it afterwards). A step-wise cost function which 
implements this goal is given by 
\begin{equation} 
g(\bx_t,u_t)=\begin{cases} -\norm{\bx_t}^2 & \text{if } \norm{\bx_t}<0.1 \\ -1 & \text{ otherwise} \end{cases}
\label{eq:pendulum_reward}
\end{equation}
Since the dimensionality of the state space is low, we can use dynamic programming (value iteration with 
grid-based interpolation) to directly determine the {\em optimal} behavioral policy, where optimal means choosing actions 
such that the accumulated costs from Eq.~\eqref{eq:pendulum_reward} are minimized among all possible 
behaviors \cite{sutton98introduction}. Comparing the results in Figure~\ref{fig:phaseplot_pendulum} (bottom row)
from using dynamic programming as opposed to using the empowerment heuristic in Figure~\ref{fig:phaseplot_pendulum} 
(top row) shows the remarkable result that with empowerment we achieve nearly the same behavior as with optimal control. 
The result is remarkable because, unlike the optimal value function, which through the underlying cost function is tied 
to a particular goal, empowerment is a generic heuristic that operates on the innate characteristics of the dynamics of the system alone.

\paragraph{Results for bicycle:} For the more complex bicycle domain, the goal is to keep the bicycle going forward by 
preventing it from falling over to one side or the other; when the angle
from the vertical axis, $\omega$, deviates 
too much from zero (that is, is greater than $\frac{12\pi}{180}$) the bicycle is considered to have fallen. Whenever 
this happens, the bicycle stops moving forward, and no matter what action the agent takes, the successor state 
will be the same for all future time steps (absorbing state), and consequently empowerment will be zero.

\begin{figure}[ht]
\begin{minipage}{0.49\textwidth}
\begin{center}
\includegraphics[width=1.0\textwidth]{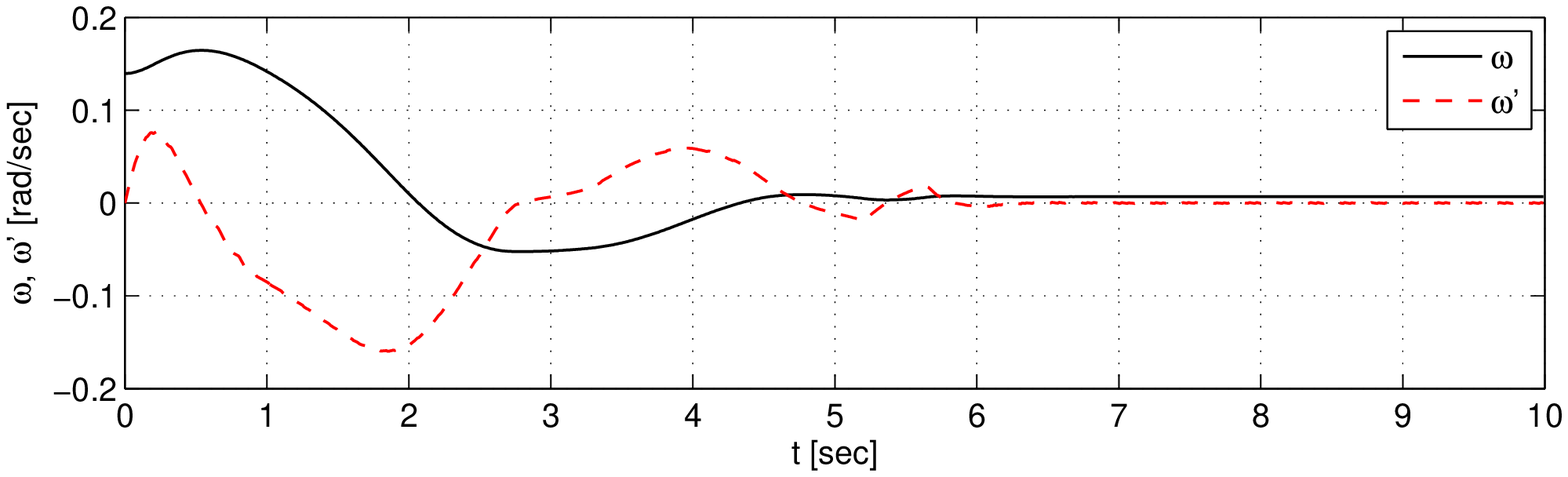}
\includegraphics[width=1.0\textwidth]{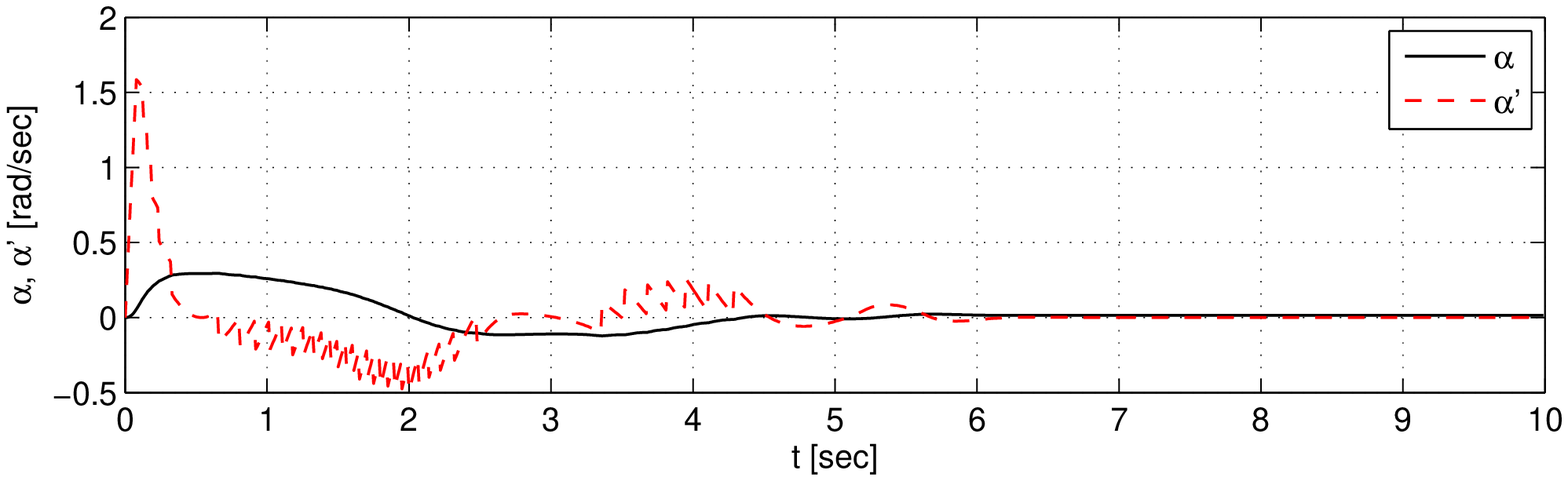}
\includegraphics[width=1.0\textwidth]{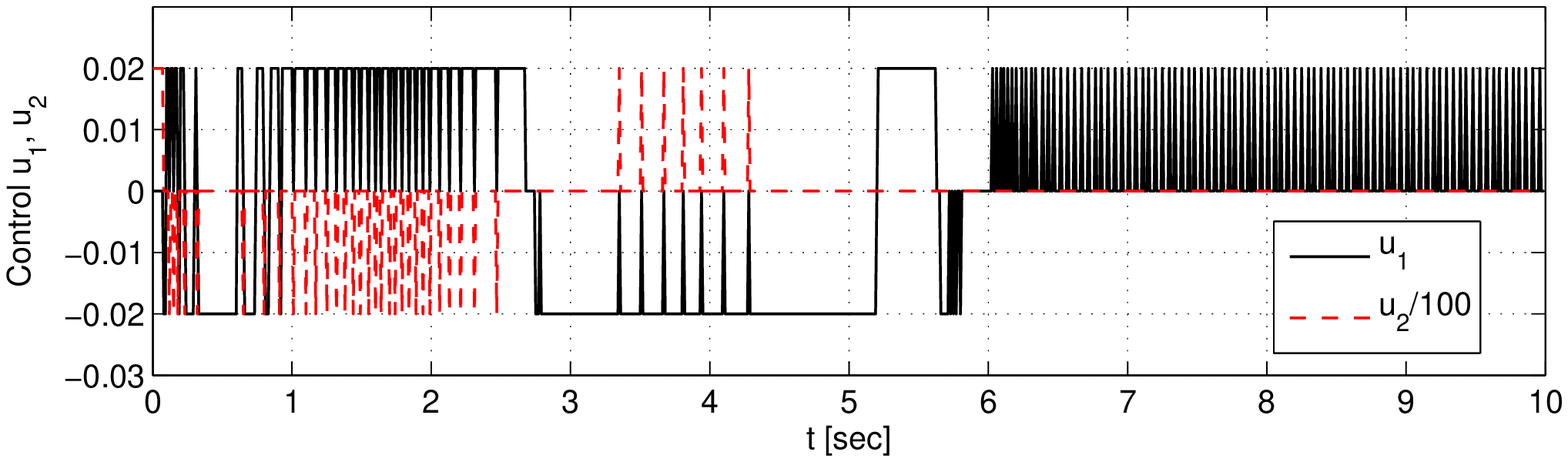}
\end{center}
\end{minipage}
\begin{minipage}{0.49\textwidth}
\begin{center}
\includegraphics[width=1.0\textwidth]{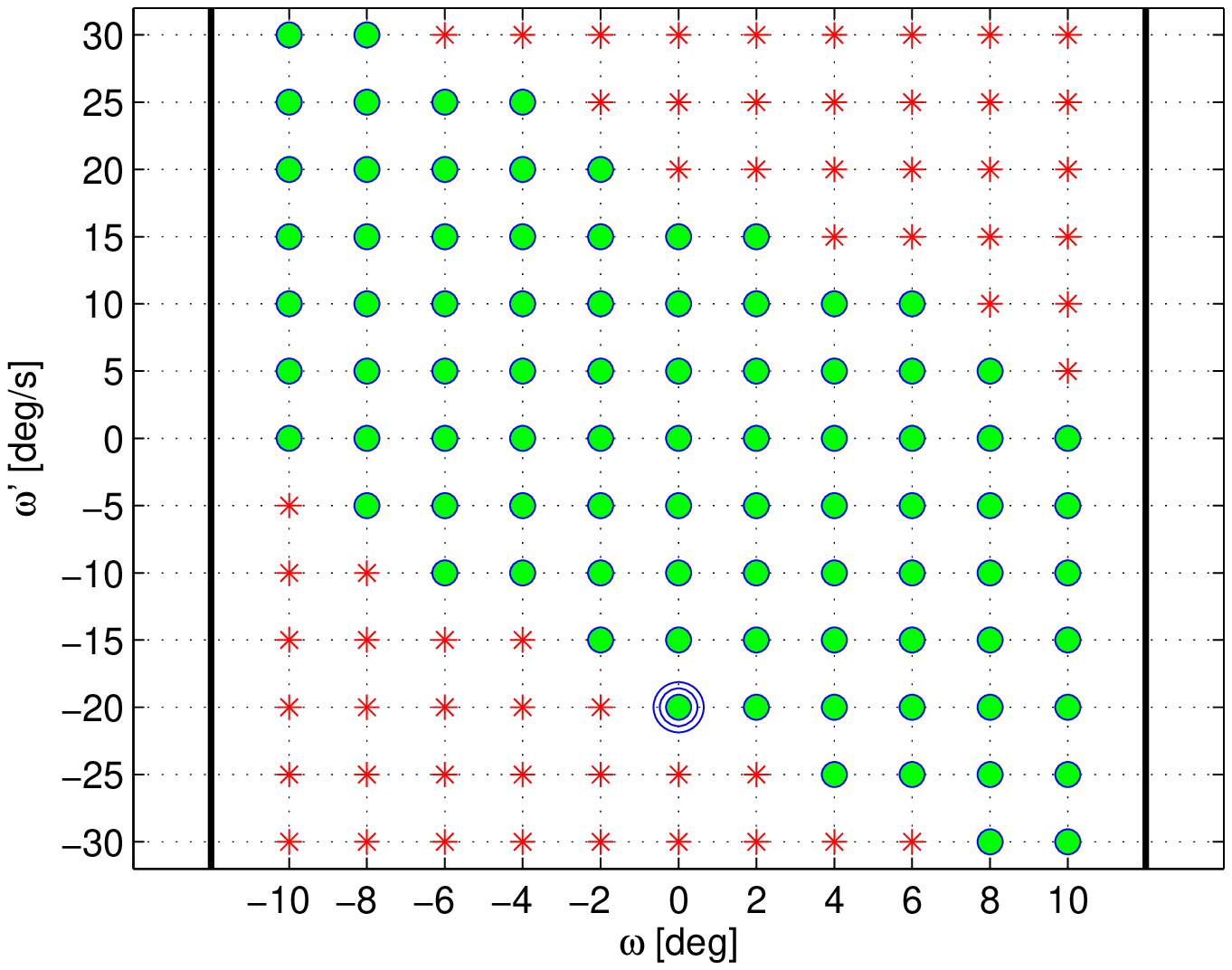}
\end{center}
\end{minipage}
\caption{Bicycle: (left side) phase plot of state variables $\omega,\dot\omega$ (upper panel), $\alpha, \dot\alpha$ (middle panel), and controls $u_1, u_2$ (lower panel) when starting from state $\bigl(\frac{8\pi}{180},0,0,0\bigr)$ and following the empowerment-based policy; (right side) shows how empowerment
is able to successfully balance the bicycle for a large variety of initial conditions; the black vertical bars indicate failure states; that is, the value of angle $\omega$ from which failure can no longer be avoided.}
\label{fig:bicycle_results}
\end{figure}

Here we examine the behavior of empowerment for different initial conditions of the bicycle: we ran different trials
by varying the angle $\omega$ in the interval 
$\frac{-10\pi}{180}, \frac{-8\pi}{180},\ldots,\frac{+8\pi}{180},\frac{+10\pi}{180}$, and $\dot\omega$ in the interval
$\frac{-30\pi}{180}, \frac{-25\pi}{180},\ldots,\frac{+25\pi}{180},\frac{+30\pi}{180}$; $\alpha$ and $\dot\alpha$ were 
initially zero in all cases. We employ $3$-step empowerment (and thus $N_n=5 \times 5 \times 5$ possible $n$-step actions) 
where each $1$-step action in an action sequence is held constant for $4$ simulation steps, and state transition noise 
$\bSigma=0.001 \mathbf{I}_{4\times 4}$. Figure~\ref{fig:bicycle_results} (right) shows that empowerment is able to keep 
the bicycle stable for a wide range of initial conditions; dots indicate that the bicycle successfully kept going 
forward for $20$ seconds, stars indicate that it did not. Note that in many cases of failure, it would actually have 
been physically impossible to prevent the bicylce from falling; for example, when the bicycle already is strongly leaning 
to the left and further has velocity pointing to the left. Also note that the column corresponding to zero angle shows an 
outlier\footnote{The outlier is a result of inaccuracy produced from Monte-Carlo approximation. Repeating the 
experiment with a larger number of samples showed that indeed the bicycle can be balanced from both initial conditions. 
However, note that these initial conditions were already close to the boundary from where balancing becomes 
impossible, regardless of how many samples are used.}; while empowerment was able to balance the bicycle 
for $\dot\omega=\frac{-20\pi}{180}$, it was not for $\dot\omega=\frac{+20\pi}{180}$. Figure~\ref{fig:bicycle_results} (left) 
shows a phase plot when starting from the initial condition $\omega=\frac{8\pi}{180}$; as we can see, empowerment keeps 
the bicycle stable and brings the system close to the point $(0,0,0,0)$, from where it can be kept stable indefinitely.

\begin{figure}[t]
\begin{center}
\hspace*{-1.25cm}\includegraphics[width=1.15\textwidth]{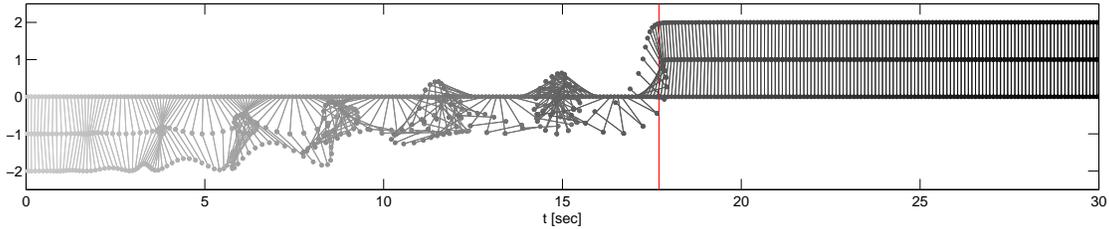}
\end{center}
\caption{Empowerment alone makes the acrobot swing up, approach the unstable equilibrium, and balance in the inverted handstand position indefinitely.}
\label{fig:acrosnapshot}
\end{figure}

\begin{figure}[htbp]
\begin{minipage}{0.49\textwidth}
\begin{center}
\includegraphics[width=0.8\textwidth]{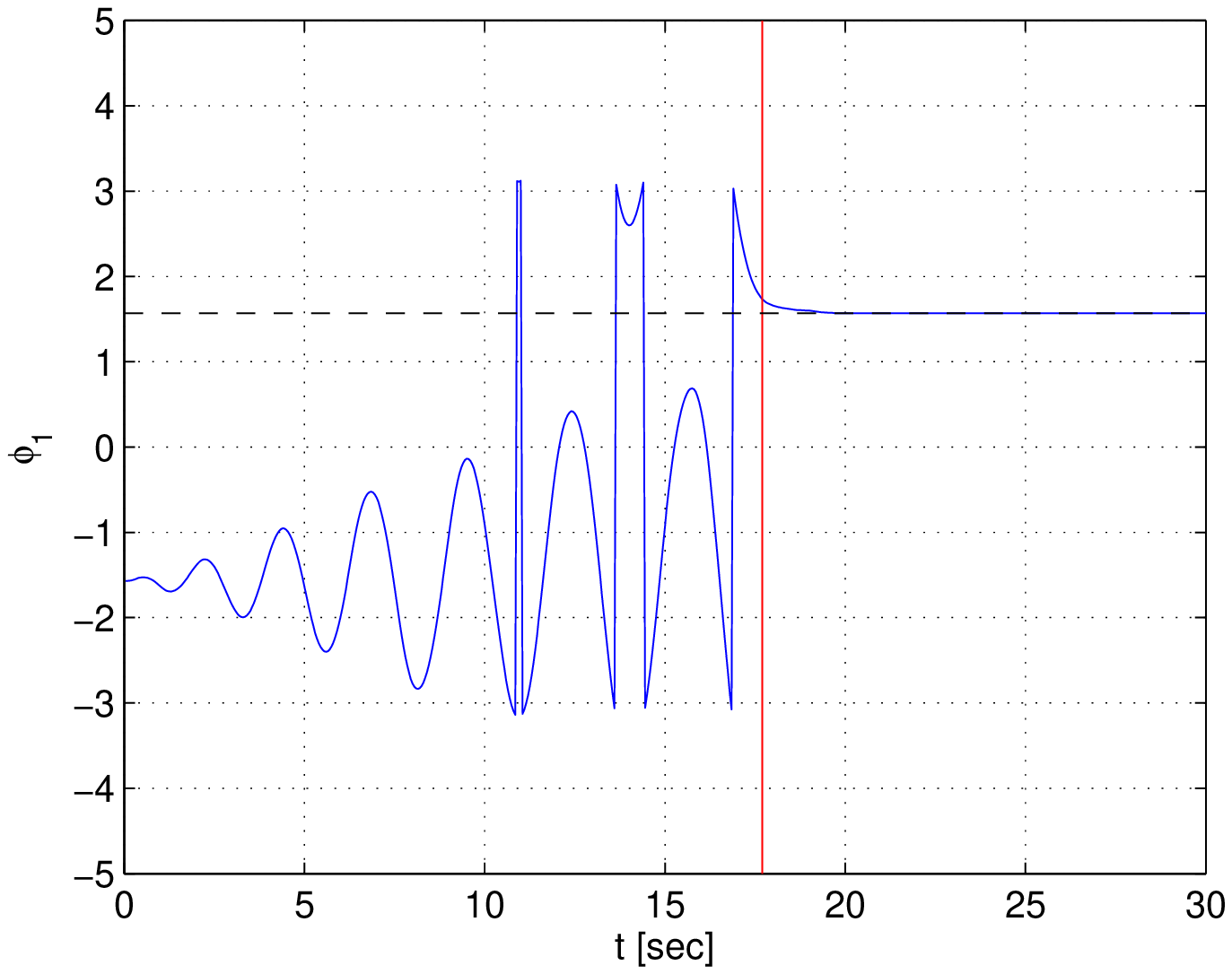}
\includegraphics[width=0.8\textwidth]{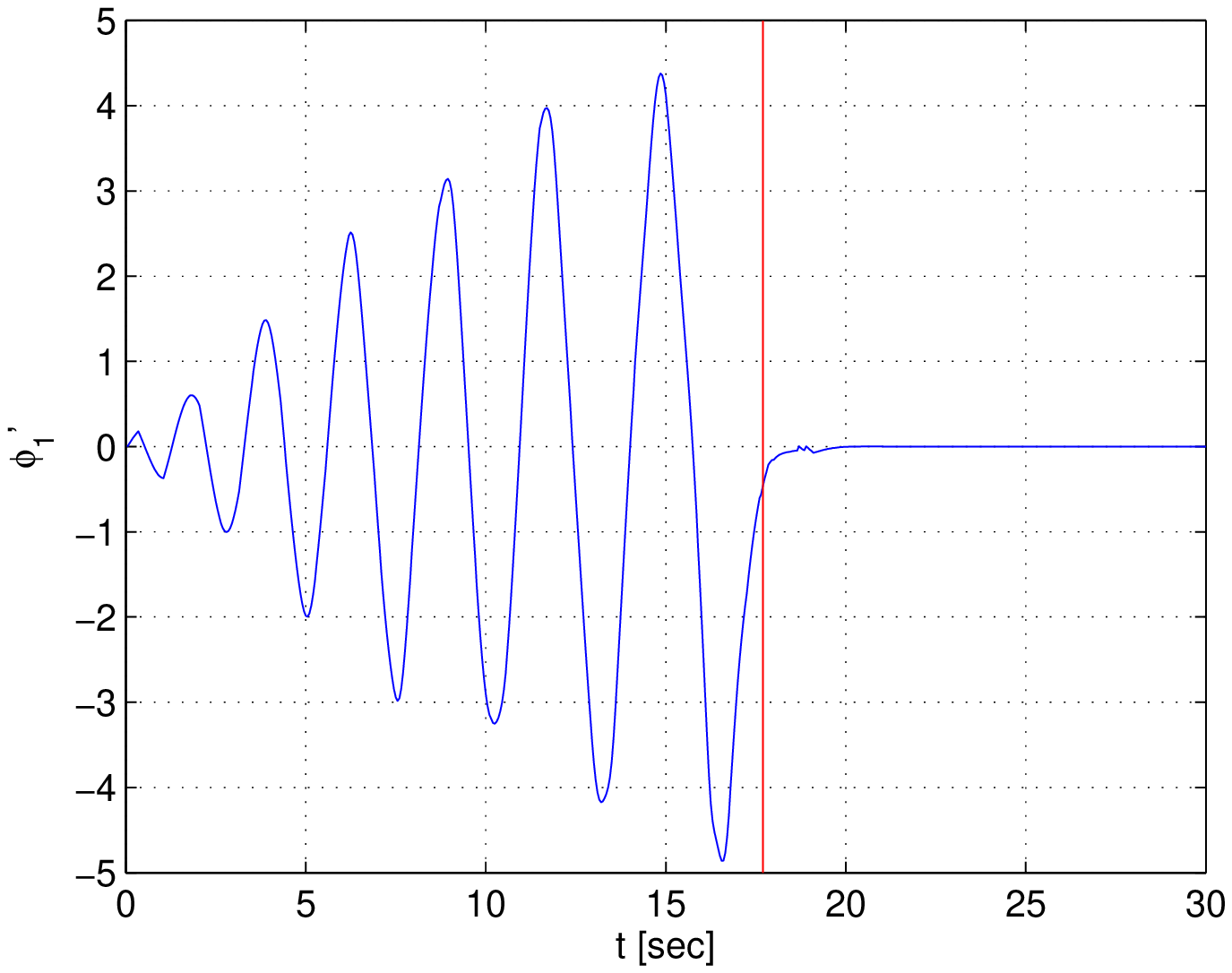}
\includegraphics[width=0.8\textwidth]{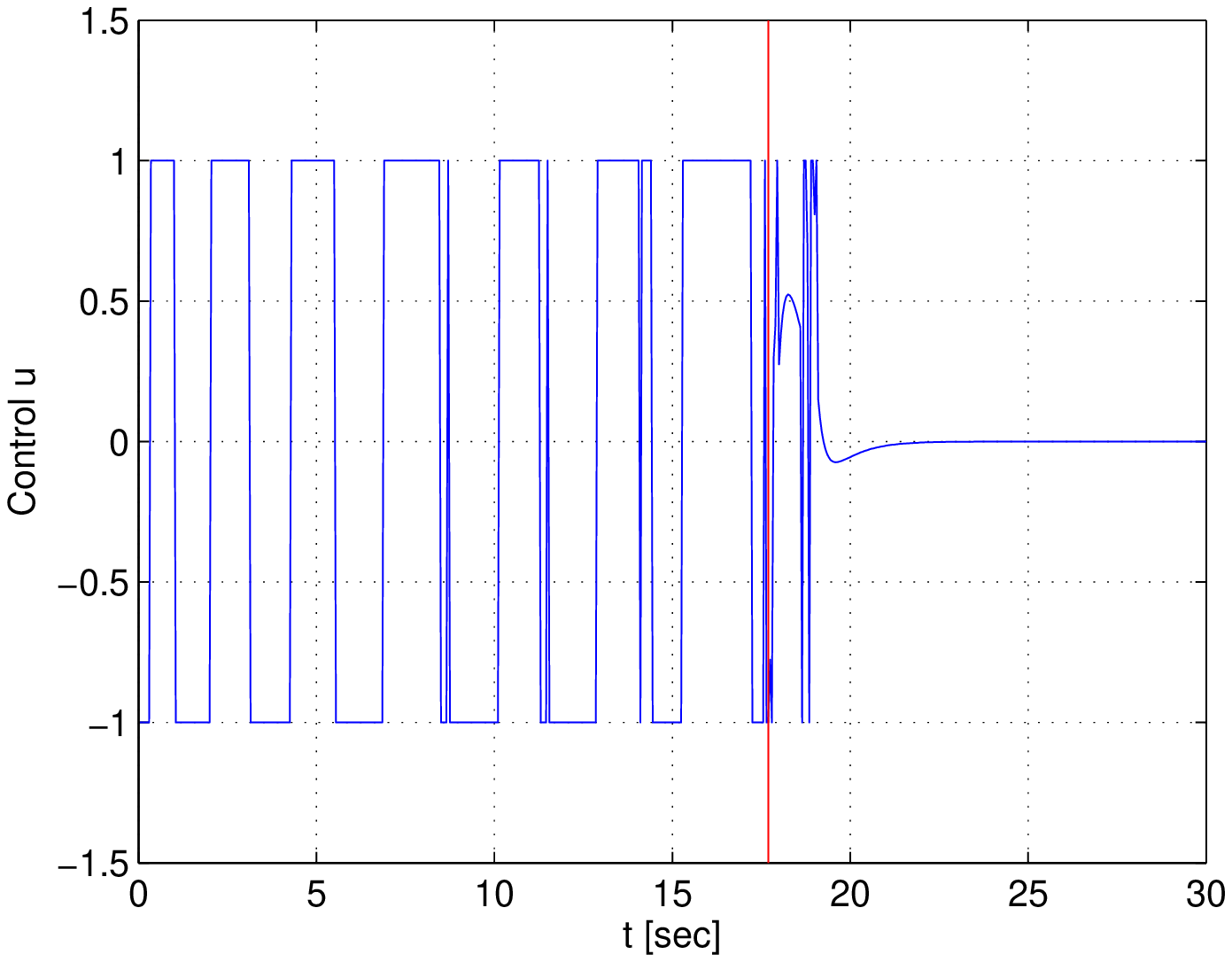}
\end{center}
\end{minipage}
\begin{minipage}{0.49\textwidth}
\begin{center}
\includegraphics[width=0.8\textwidth]{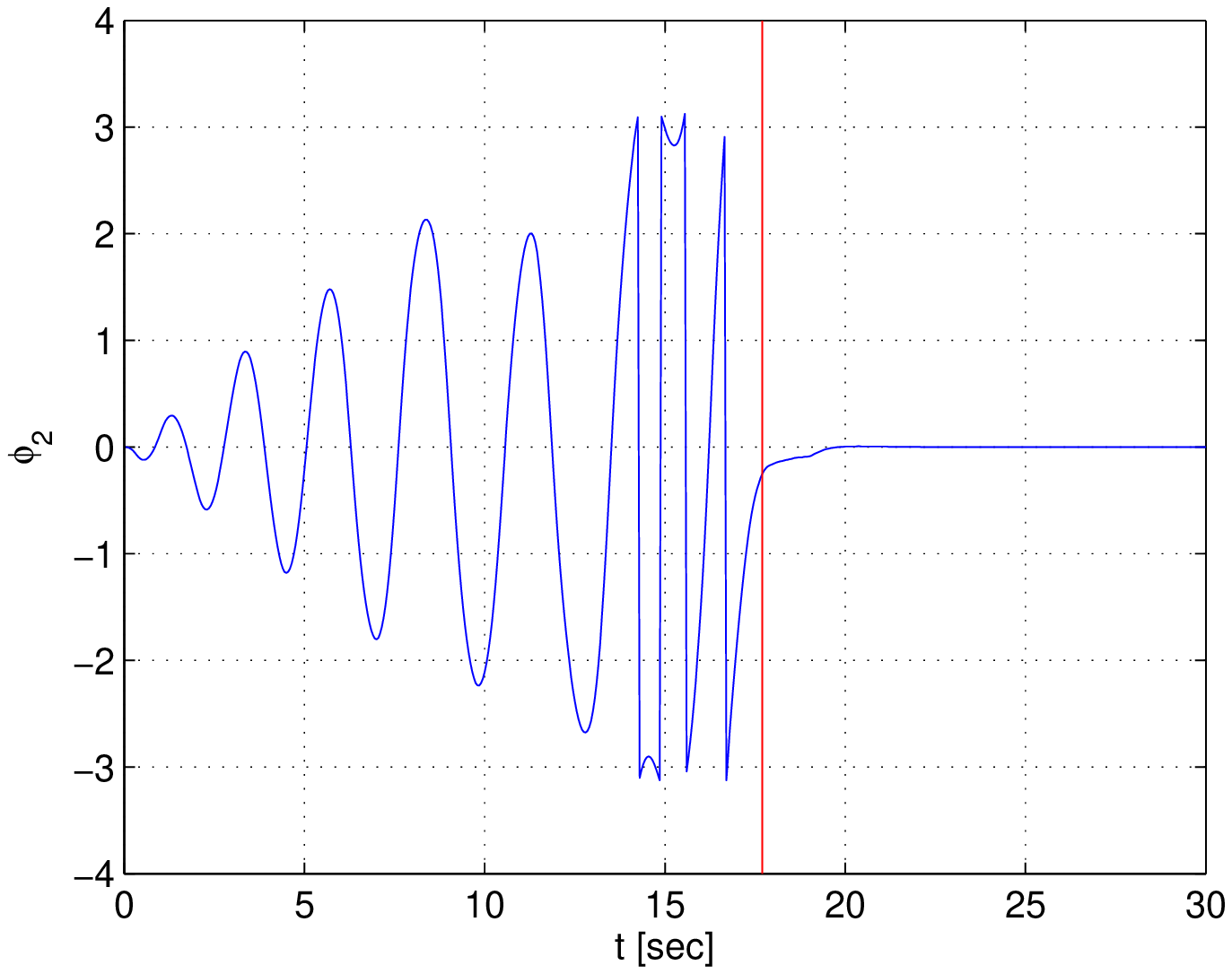}
\includegraphics[width=0.8\textwidth]{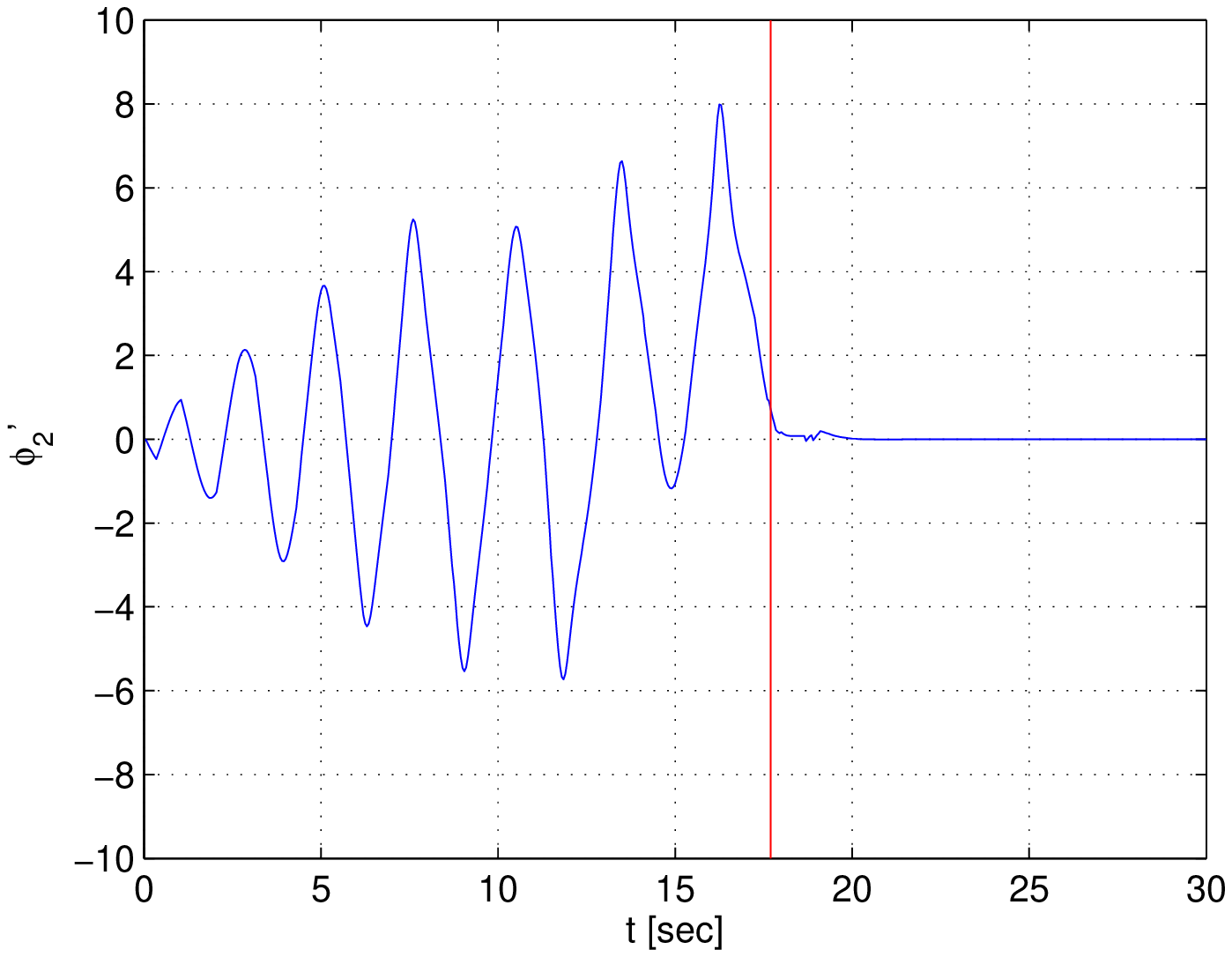}
\includegraphics[width=0.8\textwidth]{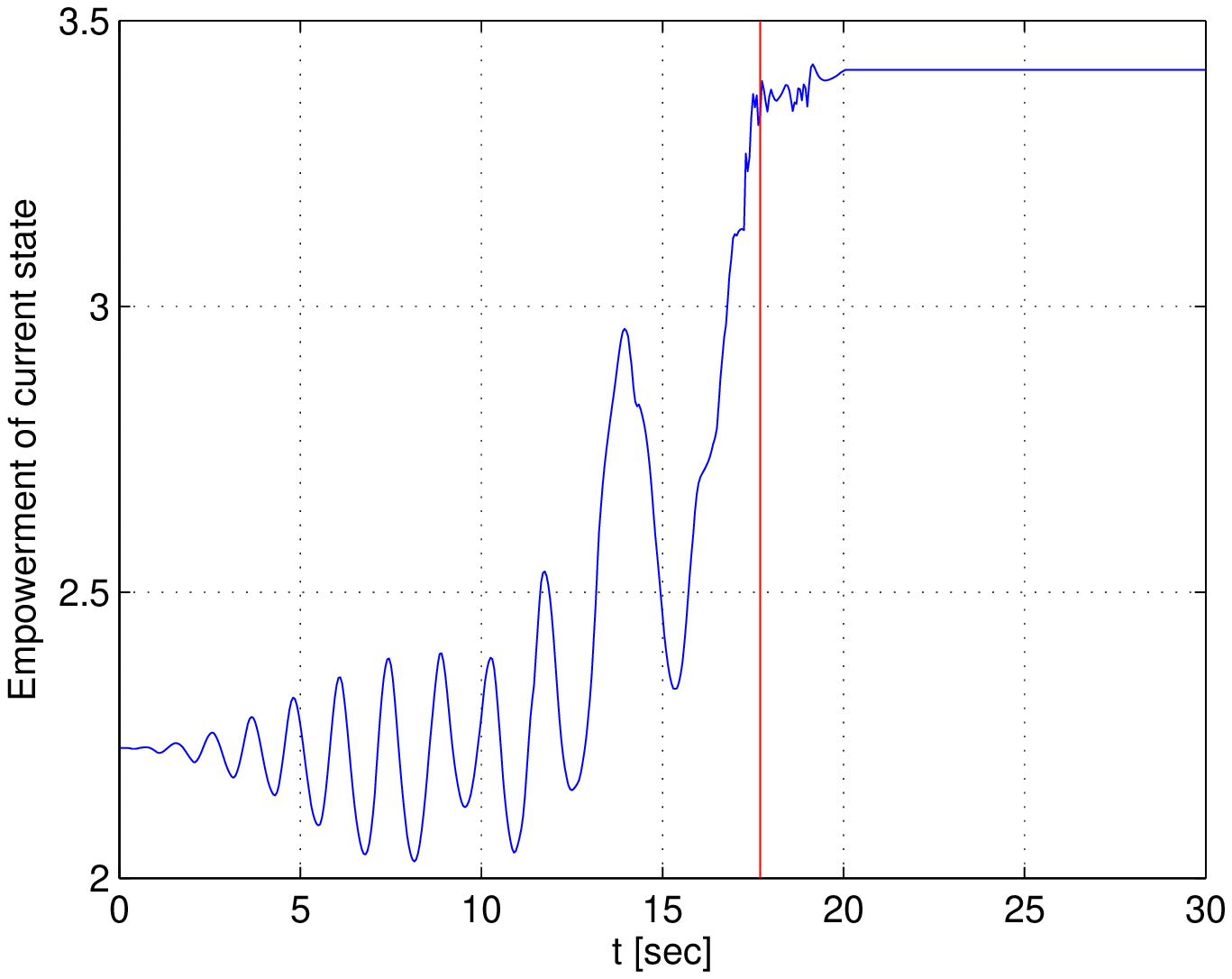}
\end{center}
\end{minipage}
\caption{Acrobot: phase plot when following the empowerment-based policy. The bottom right panel shows the associated empowerment values. The vertical bar shows the first time the
'balance' action was chosen and produced values between the extreme controls $-1$ and $+1$.}
\label{fig:acro_results}
\end{figure}

\paragraph{Results for acrobot:} For the highly challenging acrobot we require a deeper lookahead: here we consider 
$5$-step empowerment (and thus $N_n=3 \times 3 \times 3 \times 3 \times 3$ possible $n$-step actions), where each 
$1$-step action in an action sequence is held constant for $4$ simulation steps, and state transition noise 
$\bSigma=0.01 \mathbf{I}_{4\times 4}$. The phase plot in Figure~\ref{fig:acro_results} demonstrates that empowerment then leads to a 
successful swing-up behavior, approaches the unstable equilibrium, and in particular makes the agent actually balance 
in the inverted handstand position. Figure~\ref{fig:acrosnapshot} illustrates how these numbers translate into the real physical system. 
Figure~\ref{fig:acro_results} (bottom right) shows the corresponding empowerment, that is, it shows for every time step the empowerment value 
of the state the agent is in; while empowerment does not increase monotonically in every single time step, 
it increases over the time and reaches the maximum in the handstand position. The vertical bar in the figure indicates the
point where the 'balance' action was chosen for the first time as the action with highest empowerment. From this point on,   
just choosing the 'balance' would have been sufficient; however, the phase plot of the control variable reveals that during this 
phase, the balance action was not always the one with the highest empowerment.\footnote{This observation was not due to inaccuracies 
because of Monte-Carlo approximation. However, while empowerment does not exactly produce the sequence of minimal-time optimal controls,
its qualitative behavior is close.} Note that the 'balance' action (see Eq.~\eqref{eq:acro_lqr} in the appendix)
produces values in the interval $[-1,+1]$ only for states very close to the handstand position and, because of saturation, 
behaves like the two other actions $+1$ or $-1$ otherwise.

\subsection{Second scenario: model learning and exploration}
In the second experiment we will discuss a scenarion for empowerment which extends its potential applicability;
here we are interested in model learning and using empowerment to extrapolate ``intelligently'' which part of the state 
space to explore next. In particular, we will consider the case of {\em online} model learning; i.e., learning the state
transition probabilities from the samples an agent experiences while interacting with the environment (which  
is more challenging since in general we cannot generate transitions at arbitrary points in the state space and have
to make do with the states encountered during a specific -- and realistically achievable -- run). The key idea
here will be to show that with empowerment we can avoid sampling the state space exhaustively, and instead can 
learn the target behavior from only very few system-agent interactions.

\subsubsection{Overview of the learning architecture}
\label{sec:algorithm2}
\begin{figure}
\begin{center}
\psfrag{xtt}{\tiny$\mathbf x_{t+1}$}
\psfrag{xt}{\tiny$\mathbf x_{t}$}
\psfrag{at}{\tiny$a_{t}$}
\psfrag{triplet}{\tiny$(\mathbf x_t,a_t,\mathbf x_{t+1})$}
\psfrag{Mt}{\tiny$\mathcal M_t$}
\includegraphics[width=0.5\textwidth]{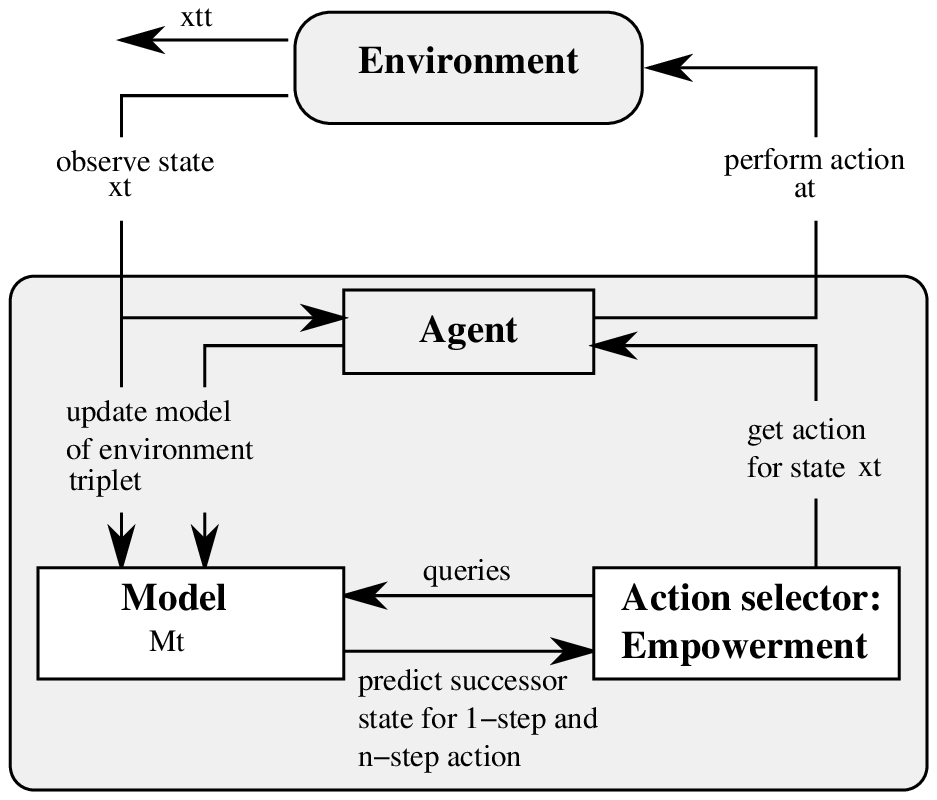}
\end{center}
\caption{A framework for model learning and empowerment-based exploration.}
\label{fig:framework}
\end{figure}

An overview of the learning architecture is depicted in Figure~\ref{fig:framework}. The agent consists of two components. One is the model learner $\mathcal M_t$, which stores a history of all transitions $\mathcal D_t=\{\bx_i,a_i,\bxx_i\}_{i=1}^t$ seen up to the current time $t$ and which implements multiple GPs to provide $1$-step predictions $p(\bx_{t+1}|\bx_t, a_t,\mathcal M_t)$ (Section~\ref{sect:Learning 1-step system dynamics}) and $n$-step predictions $p(\bx_{t+n}|\bx_t,\vec a_t^n,\mathcal M_t)$ (Section~\ref{sect:From 1-step to n-step models}). 
%
%
%
The second component is the action selector. Given the current state of the environment, we first determine the successor states under each of the possible $1$-step actions using the mean\footnote{Note that, for simplicity, we ignore that the GP-based model learner produces a predictive distribution over successor states and thus naturally incorporates noise/stochasticity of transitions. Computationally it would become quite unwieldy to calculate at every step the {\em expected} empowerment value of a successor state, as there is no closed form solution and the integral would have to be evaluated approximately, e.g. again by Monte-Carlo approximation. Note that this simplification did not change our results.} of the predictions from $\mathcal M_t$. For each successor state, we then determine their empowerment value (Section~\ref{sec:Example: Gaussian model}) using $n$-step predictions from $\mathcal M_t$. Since the predicted successor states depend on the accuracy of $\mathcal M_t$, we adjust their empowerment scores by the uncertainty of the associated $1$-step prediction. This uncertainty is taken to be the sum of the individual uncertainties of the state components in Eq.~\eqref{eq:10b}. We employ what is called {\em optimism in the face of uncertainty}: the less certain the system is, the more we want it to perform an exploratory action. Here, we linearly interpolate between the two extremes maximum uncertainty (where we assign $\log N_n$, the upper bound on empowerment) and minimum uncertainty (where we assign the actual empowerment score). 
The concrete value of the maximum uncertainty, $\beta>0$, and minimum uncertainty, $\alpha \ge0$, depend on the hyperparameters of the GPs implementing $\mathcal M_t$, for details see \cite{raswil06gp}. At the end, the agent executes the highest ranked action, observes the outcome and updates the model $\mathcal M_t$ accordingly (for performance reasons only every $K$ steps). A summary of the control loop is shown below:
\begin{enumerate}
\item {\bf Initialize:}
   \begin{enumerate}
      \item Generate initial transitions $\mathcal D_0$.
      \item Learn initial model $\mathcal M_0$.
   \end{enumerate}
\item {\bf Loop: $t=1,2,\ldots$}
  \begin{enumerate}
  \item Observe current state $\bx_t$
  \item For each $1$-step action $\nu=1,\ldots,N_a$
        \begin{enumerate}
        \item Compute $1$-step successor under $\nu$ using $\mathcal M_t$             
              (Section~\ref{sect:Learning 1-step system dynamics})               
              \[  p(\bx_{t+1}^\nu|\bx_t,a_t=\nu,\mathcal M_t)=\mathcal N
                  (\bmu_\nu(\bx_t),\bSigma_\nu(\bx_t))
              \]  
        \item Compute $n$-step empowerment $c_t^\nu:=c(\bmu_\nu(\bx_t))$
              (Section~\ref{sec:Example: Gaussian model}) using $n$-step predictions provided by
              $\mathcal M_t$ (Section~\ref{sect:From 1-step to n-step models}).
        \item Adjust empowerment scores according to the scalar uncertainty $\tr
              \bSigma_\nu(\bx_t)$ of the $1$-step prediction in $\bx_t$, linearly interpolating between
              $\log N_n$ (max uncertainty) and $c_t^\nu$ (min uncertainty):
              \[
                 \tilde c_t^\nu:=c_t^\nu + \frac{\tr                 
                 \bSigma_\nu(\bx_t)-\alpha}{\beta-\alpha} (\log N_n-c_t^\nu)
              \]
              where $\alpha$ and $\beta$ are the min and max uncertainty values of the predictions
             (depend on the hyperparameters of $\mathcal M_t$)
        \end{enumerate} 
  \item Find best action $a_t:=\argmax_{\nu=1\ldots N_a} \tilde c_t^\nu$
  \item Execute $a_t$. Observe $\bx_{t+1}$. Store transition $\mathcal D_{t+1}=\mathcal D_t \cup
        \{\bx_t,a_t,\bx_{t+1}\}$.
  \item Every $K$ steps: update model $\mathcal M_t$ using $\mathcal D_t$.      
  \end{enumerate}
\end{enumerate}

\subsubsection{Results}

For this experiment, we will only consider the inverted pendulum domain for which it will be comparatively easy, because of low
dimensionality, to compute the respective optimal behavior. The dynamics of the domain is modified to obtain an episodic
learning task: every $500$ steps, the state of the system is reset to the initial condition $(\pi,0)$, and a new episode starts. 
The action selector computes empowerment using the same parameters as in the previous section, with the difference that now 
$1$-step and $n$-step successor states are predicted by the current model. The model learner is updated (re-trained) every $K=10$ 
samples; for the GPs we employ the ARD kernel \cite{raswil06gp} with automatic selection of hyperparameters. 

For comparison, we consider RMAX \cite{RMAX-JMLR02}, a common model-based reinforcement learning 
algorithm, which also combines exploration, model learning and control, and operates not unlike the learning framework we have described 
in Section~\ref{sec:algorithm2}. The main difference is that RMAX is derived from dynamic programming and value iteration and
finds agent behavior that optimizes a given performance criterion. The performance criterion, as before, is the explicit cost function 
Eq.~\eqref{eq:pendulum_reward}, which makes the agent want to reach the goal as fast as possible. For RMAX we have to learn a model both for
the transitions of the environment and the cost function. While the former could be done with GPs (same as with empowerment), 
the latter can not be done by GPs. The reason is that the cost function is flat in every part of the state space except for a very 
small region about the goal. Since all the initial samples the agent experiences will be from the flat region, a GP would rapidly conclude
that the whole cost function is flat; since the uncertainty of the model guides exploration, the GP would predict a $-1$ cost for all states
with very high confidence, and thus the agent would miss the goal for a long time (creating a ``needle-in-a-haystack'' situation).

As it is usually done for RMAX, we therefore use a grid-based discretization to estimate costs and transitions.\footnote{The value iteration 
part of RMAX is also carried out with interpolation on a high-resolution grid. However, the details of this step are of no concern in 
this paper, and the performace comparison we make is unaffected by it.} Uncertainty of a prediction then depends on whether or not the
underlying grid-cell has been visited before. Since in RMAX unvisited states are more attractive than reaching 
the goal, the agent tends to explore the environment exhaustively before it can behave optimally. 

\begin{figure}[th]
\begin{center}
\includegraphics[width=0.9\textwidth]{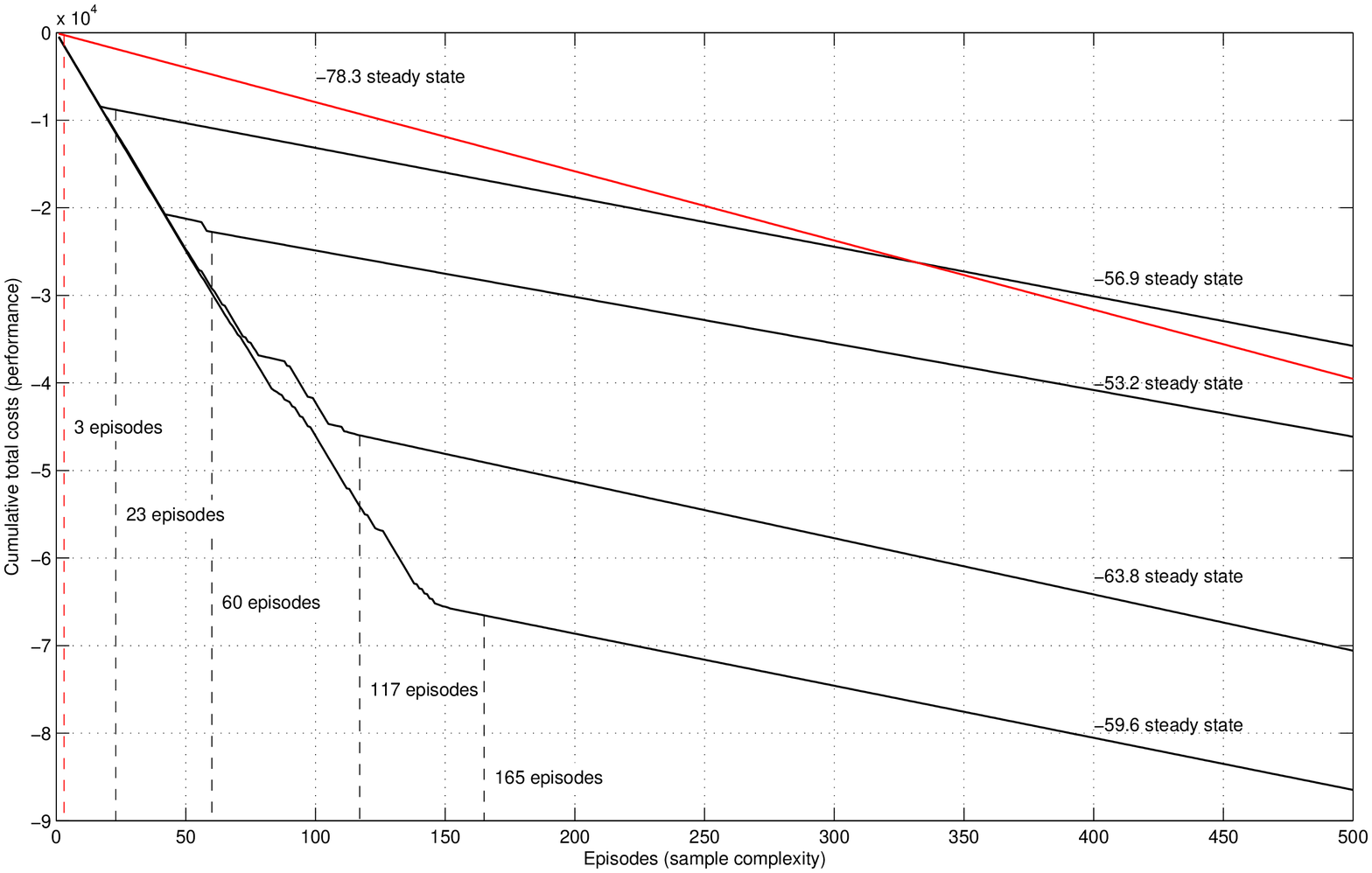}
\end{center}
\caption{Exploration and model-based learning in the inverted pendulum domain. The plot compares both the sample efficieny 
and ultimate performance of the learned behavior for empowerment with GPs (top curve) and RMAX with different levels of 
discretization: grid sizes $25$, $50$, $75$, $100$ (bottom curves).}
\label{fig:rmaxcomparison}
\end{figure}

In Figure~\ref{fig:rmaxcomparison} we compare our empowerment-based exploration with RMAX for various spacings of the underlying grid: we examine division 
into $25,50,75,100$ cells. Every curve shows the cumulative costs (under cost function Eq.~\eqref{eq:pendulum_reward}) as a function 
of episode. Thus every curve has two parts: a transient one where the agent is still learning and acting non-optimally, and a steady-state
one where the agent is acting optimally with respect to its underlying bias which is either maximizing empowerment or minimization of costs. 

The graph shows two things: (1) the finer the resolution of the grid, the longer it takes RMAX to act optimally. 
For a grid of size 25, the agent reaches optimal performance after 23 episodes; for a grid of size 50 it needs 60 
episodes; for a grid of size 75 it needs 117 episodes; and for a grid of size 100 it needs 165 episodes.  
On the other hand, empowerment only needs 3 episodes until steady-state behavior is reached. (2) The steady-state 
performance of empowerment is somewhat worse than that of RMAX, about $56$ versus $78$. However, this is not at all surprising. 
Empowerment does not at all consider the externally defined cost function when making decisions, whereas RMAX
specifically optimizes agent behavior such that performance with respect to this particular cost function is maximized.  
Still, behavior under empowerment is close to what we would achieve by explicitly optimizing a cost function; however, with 
empowerment, the agent can learn this behavior much faster since it does not have to exhaustively explore the state space
(it only has to explore the state space to the extent of learning an accurate model for state transitions).  

\begin{figure}[ht]
\begin{center}
\begin{minipage}{0.4\textwidth}
\begin{center}
\includegraphics[width=0.9\textwidth]{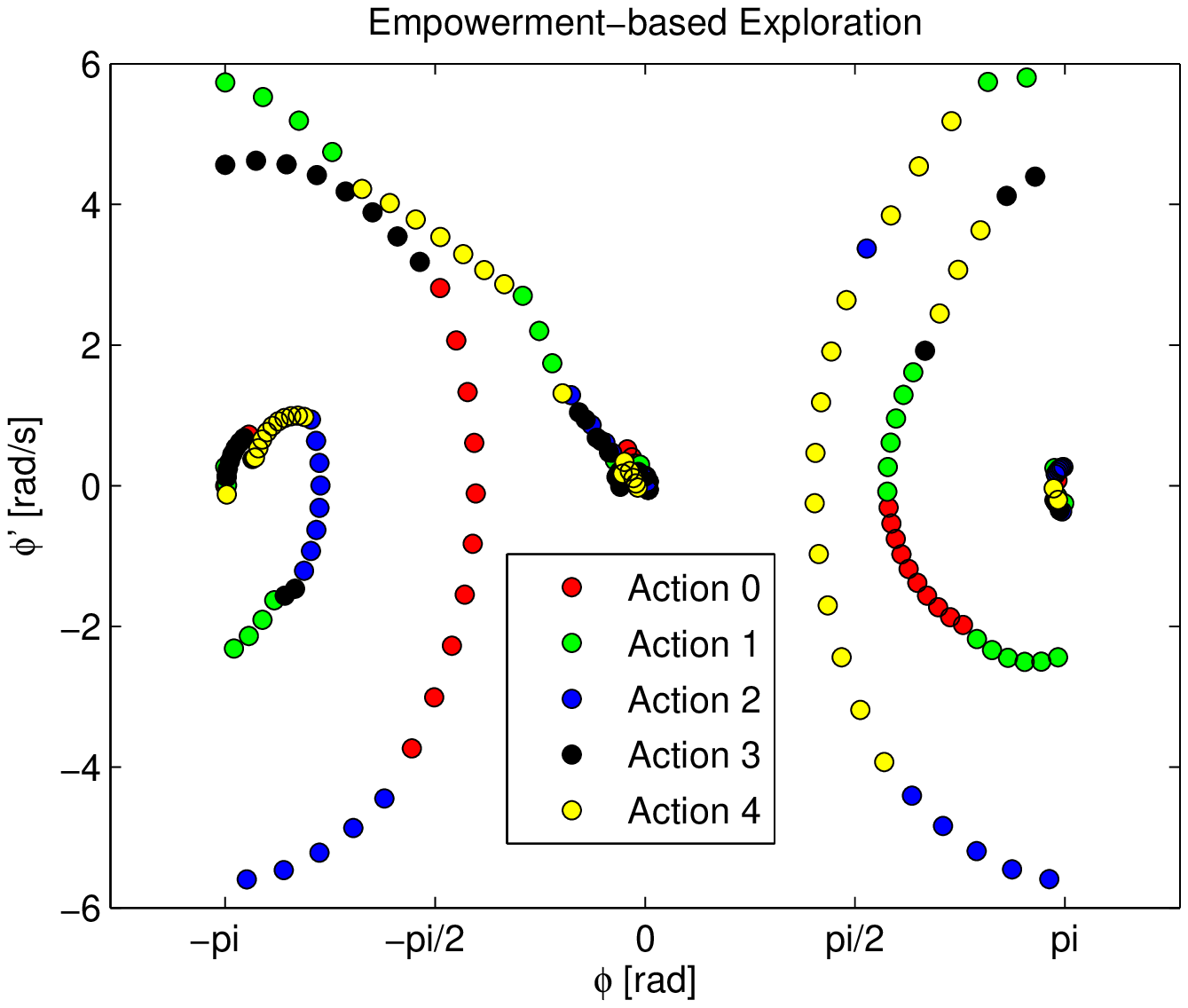}
\centerline{\small (a) Empowerment: 250 transitions}
\end{center}
\end{minipage}
\begin{minipage}{0.4\textwidth}
\begin{center}
\includegraphics[width=0.9\textwidth]{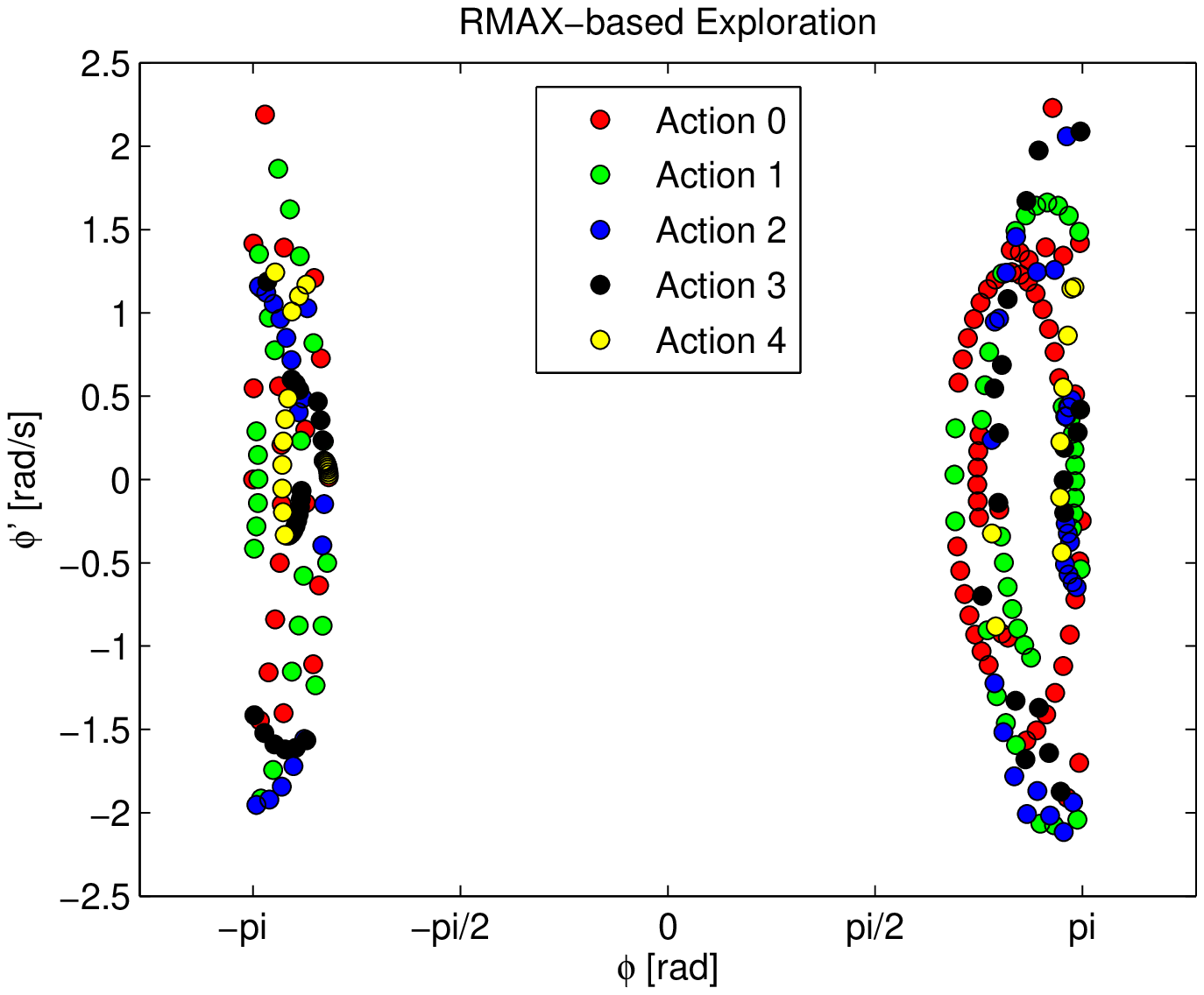}
\centerline{\small (b) RMAX: 250 transitions}
\end{center}
\end{minipage}

\bigskip
\begin{minipage}{0.4\textwidth}
\begin{center}
\includegraphics[width=0.9\textwidth]{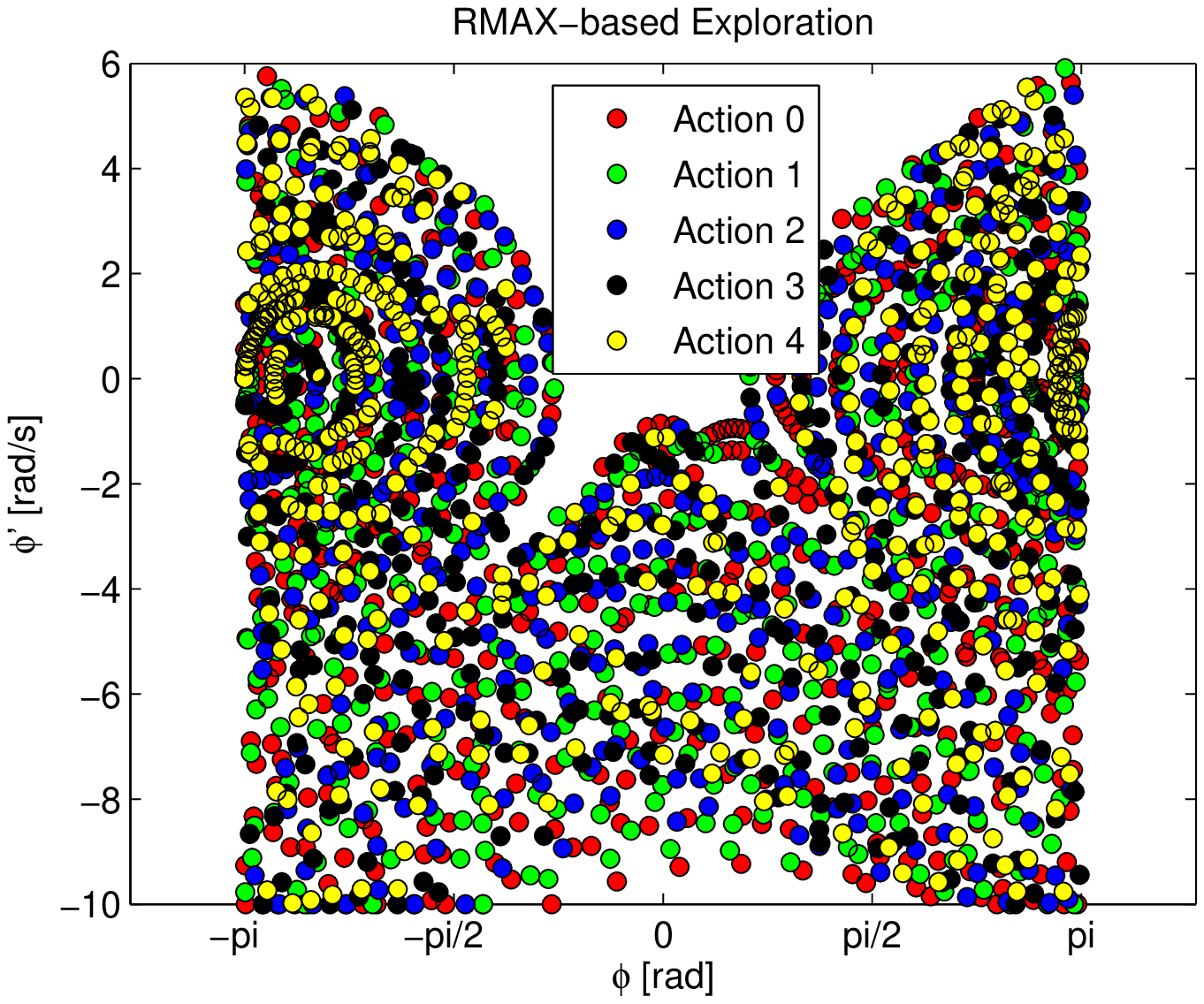}
\centerline{\small (c) RMAX: 2500 transitions}
\end{center}
\end{minipage}
\begin{minipage}{0.4\textwidth}
\begin{center}
\includegraphics[width=0.9\textwidth]{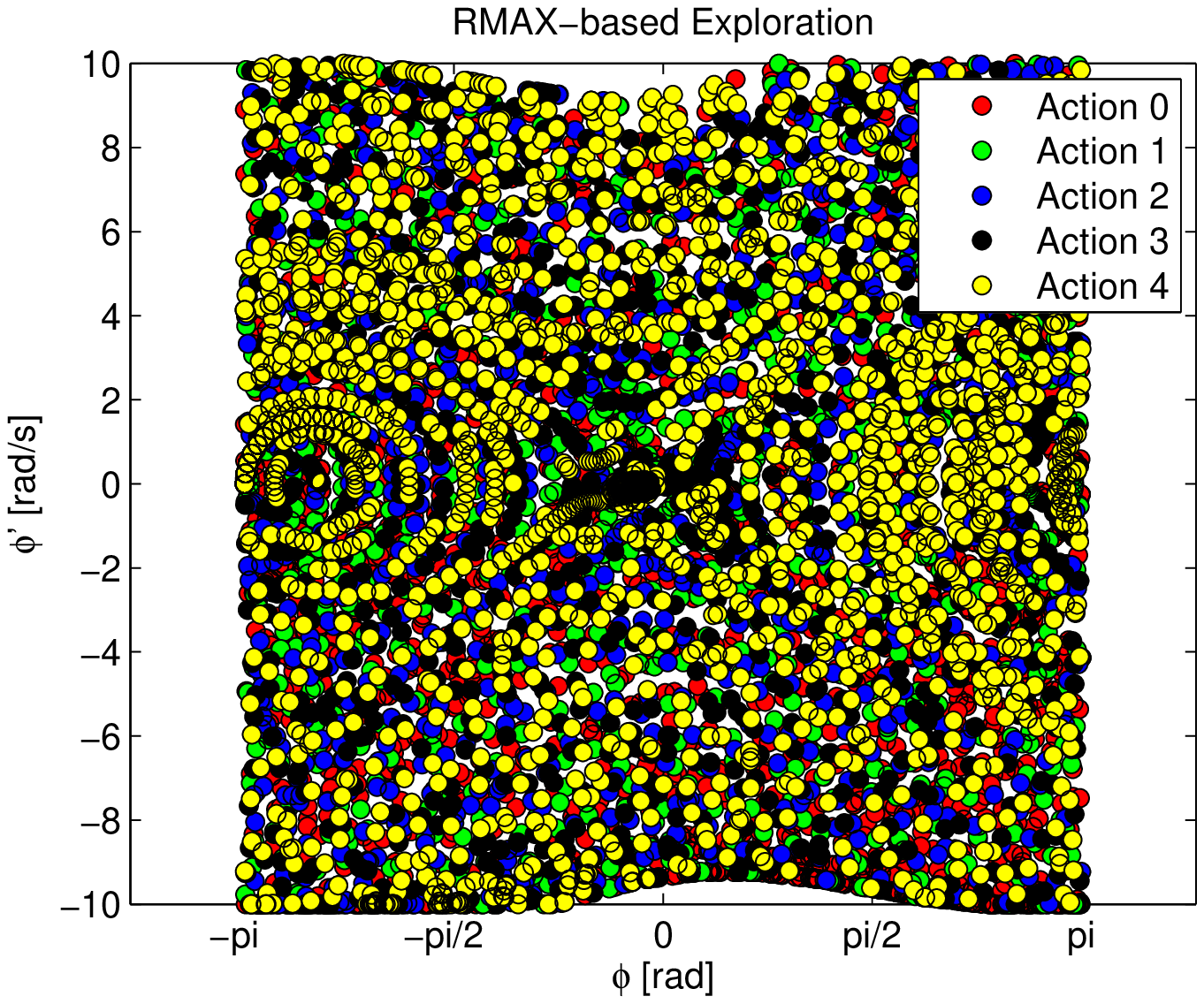}
\centerline{\small (d) RMAX: 10000 transitions}
\end{center}
\end{minipage}
\end{center}
\caption{Distribution of visited state-action pairs for empowerment and RMAX. Empowerment reaches the goal region around the point $(0,0)$ after about 250 transitions right in the very first episode, whereas RMAX needs more than ten times as long. With empowerment, the agent only has to explore limited parts of the state-action space until the model is learned. Under RMAX, in order to also learn the external cost function, the state-action space needs to be sampled exhaustively.}
\label{fig:comparison}

\end{figure}

Figure~\ref{fig:comparison} shows in more detail how empowerment drives the agent to visit only the relevant 
part of the state space. The figure compares, for empowerment and RMAX with grid spacing 25, what state-action pairs 
are visited during learning at various points in time (note that in both cases the model learner treats actions 
independently from each other and does not generalize between them). The plots show that, for the empowerment-based agent, 
the GP-based model learner can accurately predict state transitions after having seen only few very samples. As the accuracy of predictions
goes up, uncertainty of predictions goes down, as the GP becomes more confident about what it does. Low uncertainty
in turn means that the agent no longer takes exploratory actions, but instead chooses the one with the highest 
empowerment. If the learned model is accurate enough, this is as good as knowing the true transitions function and the agent
behaves accordingly (compare with model-based results in Section~\ref{sec:experiment-1}). As the plot shows, here this  
happens very soon, right within the first episode. RMAX on the other hand has to exhaustively sample the state-action
space and essentially visit every grid-cell under each action. Thus it takes much longer to even reach the goal region and 
then learn the desired behavior.

\section{Discussion}
\label{sec:discussion}
A central question that we need to address is: why does empowerment
actually carry out intuitively desirable behaviour? In previous work,
it has been shown that this property is not spurious, but actually
reappears in a number of disparate scenarios
\cite{klyubin05:_all_else_being_equal_be_empow,klyubin05:_empow,klyubin08:_keep_your_option_open,anthony08:_prefer_states_agent,anthony09:_impov_empow}.

On the other hand, one can clearly create a scenario where empowerment
will fail to match the externally imposed goal: imagine for instance
the inverted pendulum task, where the target state is some oblique
angle $\phi\not=0$, different from the upright position. Even if the
position is sustainable (we remind the reader that the task was
underactuated), that position would clearly not match the state an
empowerment maximization strategy will try to attain. Nevertheless,
the task of placing the pole in an arbitrary oblique position $\phi\not=0$
strikes one as unnatural if nothing else is specified in the task. In
other words, balancing the inverted pendulum seems to be the most
unbiased, natural task to do in that scenario.

However, of course, there are scenarios where preferred outcomes do
not naturally arise from the system dynamics. The most obvious
examples are, e.g., mazes where one needs to reach a particular goal
state. This goal state can obviously be arbitrary, and selected
independently from the actual dynamics/topology of the system. Even in
such scenarios, empowerment still mimics/approximates the
graph-theoretic notion of \emph{centrality}
\cite{anthony08:_prefer_states_agent}; this means that empowerment
maximization will place the agent (approximately) at a location in the
world from which the expected distance to a randomly specified goal
state will be minimal. In other words, it is ``the best guess'' where
the agent should place itself in expectation of a yet unknown goal,
assuming one wishes to minimize the number of steps to the
goal\footnote{We completely omit the discussion of the case when
  different actions have different costs for different states --- this
  obviously forces one to resort to the full-fledged dynamic
  programming formalism. However, this is clearly a case where the
  specification of environmental structure and dynamics are not
  sufficient for the characterization of the task and the reward
  structure needs to be explicitly specified. The issues of balancing
  explicit rewards and the information-theoretic costs of decision
  making are intricate and are discussed in detail elsewhere
  \cite{tishby10:_infor_theor_of_decis_and_action}.}.

However, the performance in our scenarios is even better than that in
that the natural goals that one would impose a priori here seem to
be anticipated by what empowerment is trying to maximize. Now, all the
considered scenarios have one thing in common: they are survival-type
scenarios. The agent aims to stay ``alive'' and to move away from
``death'' states as far as possible (we adopt here an argument that is related to
Friston's free energy model of cognition which has been brought up in
\cite{friston06,friston09:_free_energ_princ}).

What makes this particularly interesting in the context of continuous
systems which are our point of concern in the present paper is
that the smoothness of the system informs the local empowerment
gradients around the agent's state of where the most ``alive'' states
are (and many dynamical systems have this property). But even discrete
transition graphs display --- in somewhat structured scenarios like grid-worlds or
small-world networks \cite{anthony08:_prefer_states_agent} --- this
property that the attraction basins of global or good local
empowerment optima are visible from some distance. This is
particularly striking since empowerment seems to correlate well with
measures for dominating states in graphs which have been 
hand-crafted for that purpose
\cite{anthony08:_prefer_states_agent}.

Where empowerment maximization coincides with the ``natural'' optimal
control task, it computes \emph{local} gradients towards the right
direction as opposed to optimal control/dynamic programming which
implicitly require a global picture of where the goal states are. It
is an open question what properties are required from a system to
provide these relatively large attraction basins of empowerment maxima
that are visible in local empowerment gradients. This property seems
to be present in continuous environments and in environments with some
degree of globally homogeneous structures
\cite{anthony08:_prefer_states_agent}.

Different from that are, however, novel degrees of freedom which form
``gateways'' in the state space in that they are particular locations in the 
world that grant access to new subregions in the state space (implying novel ways of 
interacting with the environment) that are otherwise inaccessable from the majority of states. 
A prime example is the taxi domain from Section~\ref{sec:informal}, where the actions of picking up and
dropping off a passenger open new degrees of freedom, but only at
specific locations in the maze (another example is the ``box pushing''
scenario where an agent's empowerment increases close to a pushable
box due to the increased number of options
\cite{klyubin05:_all_else_being_equal_be_empow}). Such gateways
are usually irregular occurences in the state space and will typically 
only be detected by empowerment if they are in reach of the
action horizon. Still, intelligent action sequence extension
algorithms such as suggested in \cite{anthony09:_impov_empow} may
provide recourse and larger effective action horizons even in these
cases. However, the examples studied in this paper do not involve any
such gateways and all require only relatively short horizons by
virtue of their smooth structure. This suggests that for the
significant class of dynamic control problems empowerment may provide
a purely local exploration and behaviour heuristic which identifies
and moves towards particularly ``interesting'' areas; the present
paper furthermore demonstrates how this can be implemented in an
efficient on-line fashion.

\section{Summary}
This paper has discussed empowerment, an information-theoretic quantity that 
measures, for any agent-environment system with stochastic transitions, the extent 
to which the agent can influence the environment by its actions. While earlier 
work with empowerment has already shown its various uses in a number of 
different domains, empowerment calculation was previously limited to the 
case of small-scale and discrete domains where state transition probabilities 
were assumed to be known by the agent. The main contribution of this paper is to 
relax both assumptions. First, this paper extends calculation of empowerment 
to the case of continuous vector-valued state spaces. Second, we discuss an 
application of empowerment to exploration and online model learning where we no 
longer assume that the precise state transition probabilities are a priori 
known to the agent. Instead, the agent has to learn them through 
interacting with the environment.   

By addressing vector-valued state spaces and model learning, this paper already 
significantly advances the applicability of empowerment to real-world scenarios. 
Still, from a computational point of view, open questions remain. One question in
particular is how to best deal with continuous, vector-valued action spaces -- so 
far we assumed in this paper that the action space could be discretized. However, 
for higher dimensional action spaces (which are common in robotic applications), 
a naive discretization will soon become infeasible.

\section*{Acknowledgments}
This work has partly taken place in the Learning Agents Research Group (LARG) at
the Artificial Intelligence Laboratory, The University of Texas at Austin,
which is supported by grants from the National Science Foundation (IIS-0917122), 
ONR (N00014-09-1-0658), DARPA (FA8650-08-C-7812), 
and the Federal Highway Administration (DTFH61-07-H-00030).
This research was partially supported by the European Commission as part of the FEELIX
GROWING project (http://www.feelix-growing.org) under contract FP6
IST-045169. The views expressed in this paper are those of the authors, and
not necessarily those of the consortium.

\appendix

\section{Dynamic model of the inverted pendulum}
\label{sec:pendulum}
Refer to the schematic representation of the inverted pendulum given in Figure~\ref{fig:domains}. The state variables are the 
angle measured from the vertical axis, $\phi(t)$ [rad], and the angular velocity $\dot\phi(t)$ [rad/s].
The control variable is the torque $u(t)$ [Nm] applied, which is restricted to the interval $[-5,5]$. 
The motion of the pendulum is described by the differential equation:
\begin{equation}
\ddot\phi(t)=\frac{1}{ml^2} \Bigl(-\mu \dot\phi(t)+m g l \sin \phi(t) + u(t)\Bigr).
\label{eq:invertedpendulum}
\end{equation}
The angular velocity is restricted via saturation to the interval $\dot\phi \in [-10,10]$.
The values and meaning of the physical parameters are given in Table~\ref{tab:inverted_pendulum}.

\begin{table}
\begin{center}
\caption{Physical parameters of the inverted pendulum domain}
\label{tab:inverted_pendulum}
\begin{tabular}{llp{10cm}}
\hline
Symbol & Value & Meaning \\
\hline
$g$  & $9.81 \ [m/s^2]$ & gravitation \\
$m$  & $1 \ [kg]$ & mass of link \\
$l$  & $1 \ [m]$ & length of link \\
$\mu$ & $0.05$ & coefficient of friction\\
\hline 
\end{tabular}
\end{center}
\end{table} 

The solution to the continuous-time dynamic equation in Eq.~\eqref{eq:invertedpendulum} is obained 
using a Runge-Kutta solver.  The time step of the simulation is 0.2 sec, during which the applied control 
is kept constant. The 2-dimensional state vector is $\bx(t)=\bigl(\phi(t),\dot\phi(t))^T$, 
the scalar control variable is $u(t)$. Since our algorithm in Section~\ref{sec:Example: Gaussian model} 
allows us to compute empowerment only for a finite set of possible $1$-step actions, we discretized the 
continuous control space into $5$ discrete action choices $a\in\{-5,-2.5,0,2.5,5\}$.

\section{Dynamic model of the acrobot}
\label{sec:dynamics_acrobot}
Refer to the schematic representation of the acrobot domain 
given in Figure~\ref{fig:domains}. The state variables are the angle of the first link measured from the horizontal
axis, $\theta_1(t)$ [rad], the angular velocity $\dot\theta_1(t)$ [rad/s], the angle between the second 
link and the first link $\theta_2(t)$ [rad], and its angular velocity $\dot\theta_2(t)$ [rad/s].
The control variable is the torque $\tau(t)$ [Nm] applied at the second joint. 
The dynamic model of the acrobot system is \cite{Spong95}:
\begin{align}
\ddot\theta_1(t)=&-\frac{1}{d_1(t)} \bigl( d_2(t)\ddot\theta_2(t)+\phi_1(t) \bigr) \label{eq:acro1}\\
\ddot\theta_2(t)=&\frac{1}{m_2l_{c2}^2+I_2-\frac{d_2(t)^2}{d_1(t)}} \Bigl(\tau(t)+
\frac{d_2(t)}{d_1(t)}\phi_1(t)-m_2 l_1 l_{c2}\dot\theta_1(t)^2 \sin \theta_2(t) - \phi_2(t) \Bigr) \label{eq:acro2}
\end{align}
where
\begin{align*}
d_1(t):=& m_1 l_{c1}^2 +m_2\bigl(l_1^2+l_{c2}^2+2 l_1 l_{c2} \cos \theta_2(t)\bigr)+I_1+I_2 \\
d_2(t):=& m_2 \bigl(l_{c2}^2+l_1 l_{c2}\cos \theta_2(t) \bigr) +I_2 \\
\phi_1(t):=& -m_2 l_1 l_{c2} \dot\theta_2(t)^2 \sin \theta_2(t) - 2 m_2 l_1 l_{c2} \dot\theta_2(t)\dot\theta_1(t)\sin \theta_2(t)
+\bigl(m_1 l_{c1}+m_2 l_1 \bigr) g \cos \theta_1(t) + \phi_2(t) \\
\phi_2(t):=& m_2 l_{c2} g \cos \bigl(\theta_1(t) + \theta_2(t)\bigr).
\end{align*}
The angular velocities are restricted via saturation to the interval $\theta_1 \in [-4\pi,4\pi]$,
and $\theta_2 \in [-9\pi,9\pi]$. The values and meaning of the physical parameters are given in 
Table~\ref{tab:acro}; we used the same parameters as in \cite{sutton98introduction}.

\begin{table}
\begin{center}
\caption{Physical parameters of the acrobot domain}
\label{tab:acro}
\begin{tabular}{llp{10cm}}
\hline
Symbol & Value & Meaning \\
\hline
$g$   & $9.8 \ [m/s^2]$ & gravitation \\
$m_i$ & $1 \ [kg]$ & mass of link $i$ \\
$l_i$ & $1 \ [m]$ & length of link $i$ \\
$l_{ci}$ & $0.5 \ [m]$ & length to center of mass of link $i$ \\
$I_i$ & $ 1 \ [kg\cdot m^2]$ & moment of inertia of link $i$ \\
\hline 
\end{tabular}
\end{center}
\end{table} 

The solution to the continuous-time dynamic equations in Eqs.~\eqref{eq:acro1}-\eqref{eq:acro2} is obained 
using a Runge-Kutta solver.  The time step of the simulation is 0.2 sec, during which the applied control 
is kept constant. The 4-dimensional state vector is $\bx(t)=\bigl(\theta_1(t),\theta_2(t),\dot\theta_1(t),\dot\theta_2(t)\bigr)^T$, 
the scalar control variable is $\tau(t)$.

The motor was allowed to produce torques $\tau$ in the range $[-1,1]$. Since our algorithm in 
Section~\ref{sec:Example: Gaussian model} allows us to compute empowerment only for a finite set 
of possible $1$-step actions, we discretized the continuous control space. Here we use three actions: the
first two correspond to a bang-bang control and take on the extreme values $-1$ and $+1$. However, 
a bang-bang control alone does not allow us to keep the acrobot in the inverted handstand position, 
which is an unstable equilibrium.
As a third action, we therefore introduce a more complex balance-action, which is derived via LQR. First, 
we linearize the acrobot's equation of motion about the unstable equilibrium $(-\pi/2,0,0,0)$, 
yielding:
\begin{equation*}
\dot \bx(t)=\mathbf A \bx(t) + \mathbf B \bu(t),
\end{equation*} 
where, after plugging in the physical parameters of Table~\ref{tab:acro},
\begin{equation*}
\mathbf A=\begin{bmatrix} 0 & 0 & 1 & 0 \\ 0 & 0 & 0 & 1 \\ 6.21 & -0.95 & 0& 0\\ -4.78 & 5.25 & 0 &0\end{bmatrix},
\quad
\mathbf B=\begin{bmatrix} 0 \\ 0 \\ -0.68 \\ 1.75 \end{bmatrix},
\quad
\bx(t)=\begin{bmatrix} \theta_1(t)-\pi/2 \\ \theta_2(t) \\ \dot\theta_1(t) \\ \dot\theta_2(t) \end{bmatrix}
\quad
\bu(t)=\tau(t).
\end{equation*}
Using MATLAB, an LQR controller was then computed for the cost matrices $\mathbf Q=\mathbf I_{4\times 4}$ and $\mathbf R=1$,
yielding the state feedback law 
\begin{equation}
\label{eq:acro_lqr} 
\bu(t)=-\mathbf K\bx(t), 
\end{equation} 
with constant gain matrix $\mathbf K=[-189.28, -47.46, -89.38,-29.19]$. The values resulting from 
Eq.~\eqref{eq:acro_lqr} were truncated to stay inside the valid range $[-1,1]$. Note that the LQR
controller works as intended and produces meaningful results only when the state is already
in a close neighborhood of the handstand state; in particular, it is incapable of swinging up 
and balancing the acrobot on its own from the initial state $(0,0,0,0)$.

\section{Dynamic model of the bicycle}
\label{sec:dynamics_bicycle}
Refer to the schematic representation of the bicycle domain 
given in Figure~\ref{fig:domains}. The state variables are the roll angle of the bicycle measured from the vertical
axis, $\omega(t)$ [rad], the roll rate $\dot\omega(t)$ [rad/s], the angle of the handlebar $\alpha(t)$
[rad] (measured from the longitudal axis of the bicycle), and its angular velocity $\dot\alpha(t)$ [rad/s].
The control variables are the displacement $\delta(t)$ [m] of the bicycle-rider common center of mass
perpendicular to the plane of the bicycle, and the torque $\tau(t)$ [Nm] applied to the
handlebar. The dynamic model of the bicycle system is \cite{Ernst05}:
\begin{align}
\ddot\omega(t)  = & \frac{1}{I_{bc}} \Bigl\{ \sin(\beta(t))(M_c+M_r)gh \nonumber \\
& -cos(\beta(t))\Bigl[\frac{I_{dc}v}{r}\dot\alpha(t)+\mathrm{sign}(\alpha(t))v^2
\Bigl(\frac{M_d r}{l}\bigl(|\sin(\alpha(t))|+|\tan(\alpha(t))|\bigr)
+\frac{(M_c+M_r)h}{r_{CM}(t)} \Bigr) \Bigr] \Bigr\} \label{eq:bic1} \\
\ddot\alpha(t) =& \begin{cases} \frac{1}{I_{dl}} \bigl(\tau(t)-\frac{I_{dv}}{r}\dot\omega(t) \bigr)
& \text{if } |\alpha(t)| \le \frac{80\pi}{180} \\ 0 & \text{otherwise} \end{cases} \label{eq:bic2}
\end{align}  
where 
\begin{equation*}
\beta(t):=\omega(t)+atan \frac{\delta(t)+\omega(t)}{h}, \qquad 
\frac{1}{r_{CM}(t)}:= \begin{cases}
\frac{1}{\sqrt{(l-c)^2+\frac{l^2}{\sin^2(\alpha(t)^2)}}} & \text{if } \alpha(t)\ne 0 \\
0 & \text{otherwise} \end{cases}.
\end{equation*}
The steering angle $\alpha$ is restricted to the interval $[\frac{-80\pi}{180},\frac{80\pi}{180}]$, 
and whenever this bound is reached the angular velocity $\dot \alpha$ is set to $0$. 
The moments of inertia are computed as:   
\begin{align*}
I_{bc}&=\frac{13}{3} M_c h^2 + M_r(h+d_{CM})^2 & I_{dc}&=M_d r^2 \\
I_{dv}&=\frac{3}{2}M_d r^2 & I_{dl}&=\frac{1}{2}M_d^2
\end{align*}
The values and meaning of the remaining physical parameters are given in Table~\ref{tab:bic}.

\begin{table}
\begin{center}
\caption{Physical parameters of the bicycle domain}
\label{tab:bic}
\begin{tabular}{llp{10cm}}
\hline
Symbol & Value & Meaning \\
\hline
$g$ & $9.81 \ [m/s^2]$ & gravitation \\
$v$ & $10/3.6 \ [m/s]$ & constant speed of the bicycle \\
$h$ & $0.94 \ [m]$ & height from ground of the common bicycle-rider center of mass\\
$l$ & $1.11 \ [m]$ & distance between front and back tire at the point where they touch the ground\\
$r$ & $0.34 \ [m]$ & radius of a tire \\
$d_{CM}$ & $0.3 \ [m]$ & vertical distance between the bicycle's and rider's center of mass\\
$c$ & $0.66 \ [m]$ & horizontal distance between front tire and common center of mass\\
$M_c$ & $15 \ [kg]$ & mass of the bicycle \\
$M_d$ & $1.7 \ [kg]$ & mass of a tire\\
$M_r$ & $60 \ [kg]$ & mass of the rider\\
\hline 
\end{tabular}
\end{center}
\end{table}

Roll rate $\dot\omega$ and angular velocity 
$\dot \alpha$ are kept in the interval $[-2\pi,2\pi]$ via saturation; roll angle $\omega$ is 
restricted to $[\frac{-12\pi}{180},\frac{12\pi}{180}]$.
Whenever the roll angle is larger than $\frac{12\pi}{180}$ in either direction, the bicycle is supposed to have fallen. 
This state is treated as a terminal state by defining all outgoing transitions as self-transitions, that is, once a 
terminal state is reached, the system stays there indefinitely, no matter what control is performed. 
Thus, to keep the bicycle going forward, the bicycle has to be prevented from falling.

The solution to the continuous-time dynamic equations in Eqs.~\eqref{eq:bic1}-\eqref{eq:bic2} is obained using a Runge-Kutta solver. 
The time step of the simulation is 0.2 sec, during which the applied control is kept constant. The 4-dimensional state vector is 
$\bx(t)=\bigl(\omega(t),\dot\omega(t),\alpha(t),\dot\alpha(t)\bigr)^T$, the 2-dimensional control
vector is $\bu(t)=\bigl(\delta(t),u(t)\bigr)^T$. Control variable $\delta$ was allowed to vary in $[-0.02,0.02]$, $\alpha$ was allowed 
to vary in $[-2,2]$. Since our algorithm in Section~\ref{sec:Example: Gaussian model} 
allows us to compute empowerment only for a finite set of possible $1$-step actions, we discretized the continuous 
control space. As in \cite{Lagoudakis03}, we only consider the following 5 discrete actions: $a_1=(-0.02,0),a_2=(0,0),a_3=(0.02,0),a_4=(0,-2),a_5=(0,2)$.

\bibliographystyle{apacite}

\end{document}